\documentclass[]{fairmeta}

\usepackage{microtype}
\usepackage{graphicx}
\usepackage{subcaption}
\usepackage{enumitem}
\usepackage{bbm}
\usepackage{multirow}
\usepackage{wrapfig}
\usepackage{amssymb}

\usepackage{booktabs} %

\usepackage{xcolor}

\usepackage{amssymb}
\usepackage{mathtools}
\usepackage{amsthm}

\usepackage{color-edits}
\addauthor{gn}{magenta}
\addauthor{sw}{cyan}
\addauthor{vv}{teal}
\addauthor{ym}{red}
\addauthor{sw}{purple}
\addauthor{dh}{green}
\addauthor{cn}{orange}

\theoremstyle{plain}
\newtheorem{theorem}{Theorem}[section]
\newtheorem{proposition}[theorem]{Proposition}

\theoremstyle{definition}

\theoremstyle{remark}
\newtheorem{remark}[theorem]{Remark}

\usepackage{microtype}
\usepackage{amsmath}

\newcommand{\argmax}{\operatornamewithlimits{argmax}}
\DeclareMathOperator{\sign}{sgn}

\title{Discretizing Reward Models}

\author[1,*]{Vijay Viswanathan}
\author[2]{Shiqi Wang}
\author[2]{Devamanyu Hazarika}
\author[2]{Chirag Nagpal}
\author[1]{Tongshuang Wu}
\author[1]{Graham Neubig}
\author[2]{Yuning Mao}

\affiliation[1]{Carnegie Mellon University}
\affiliation[2]{Meta Superintelligence Labs}

\contribution[*]{Work done at Meta}

\abstract{Despite their widespread use, the role of \emph{reward models} in shaping reinforcement learning is poorly understood. Reward models offer a tempting promise: they automatically estimate response quality in the absence of verifiers or human judges. Unlike ``verifiable rewards'' which typically produce binary scores, reward models typically produce continuous scores, allowing them to be sensitive to fine-grained differences in responses. However, we show this apparent strength is a serious weakness: many popular reward models are \textit{oversensitive}, assigning different scores to equally good responses. Theoretically, we show that seemingly perfect reward models can be highly oversensitive; empirically, this oversensitivity can lead to bad policies. In place of existing notions of ``reward model accuracy,'' we propose evaluating reward models using distinct measures of ``discriminative ability'' and ``specificity'' (the complement of oversensitivity). As a solution, we describe a training-free algorithm that uses Monte Carlo dropout on any neural reward model to produce discrete reward clusters. Theoretically, we prove there exist discretizations that reduce oversensitivity at minimal expense of discriminative ability; empirically we show, in both controlled and natural RL settings, that discretizing rewards leads to less reward hacking and better policies than training on the original rewards.
}

\date{\today}
\correspondence{Vijay Viswanathan at \email{vijayv@cs.cmu.edu}}

\begin{document}

\maketitle

\section{Introduction}
Reinforcement learning's central characteristic is the use of \emph{rewards} to rate model behavior instead of explicit demonstrations. Reinforcement learning has proven most reliable on \emph{verifiable} problems, where a program can grade any response \citep{DeepSeekAI2025DeepSeekR1IR, Lambert2024TLU3P, lin2025learningsolveverifyselfplay}. For harder-to-verify tasks, standard practice is to use reward models (``RM's'') that evaluate response quality  by prompting a pretrained LM \citep{SaadFalcon2025ShrinkingTG, Tunstall2023ZephyrDD, Sun2024RethinkingBM, RLCF} or via explicit training \citep{10.5555/3294996.3295184, Liu2024SkyworkRewardBO, ArmoRM}. Continuous-valued RM's induce a strict total order over responses. This offers the potential of capturing nuances that binary verification cannot --- rewarding partial progress  (e.g. distinguishing ``good'' from ``great'') and stylistic merit in ways that a binary pass/fail signal cannot. \textbf{We argue this apparent strength is a serious weakness, due to \emph{reward model oversensitivity}.}

\begin{figure}[t]
  \centering
  \includegraphics[width=\linewidth]{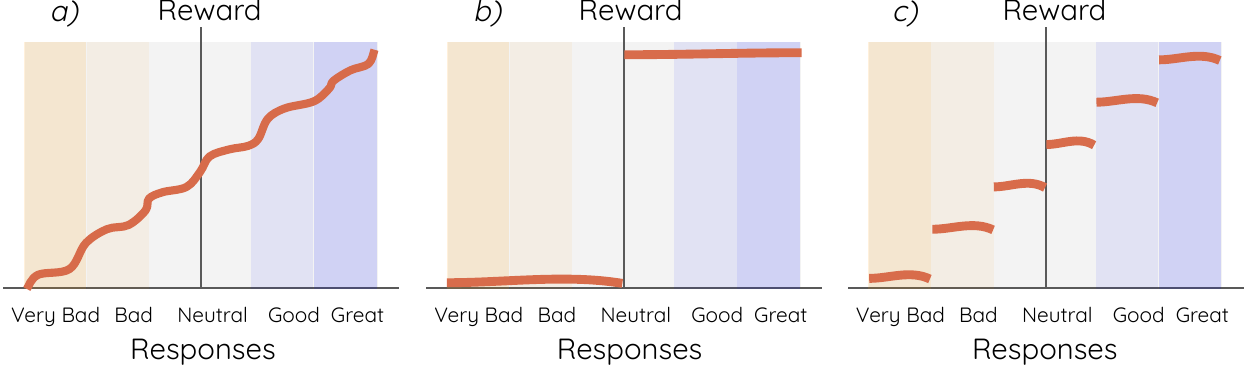}
  \caption{Reward models can be strong at discriminating ``good responses'' from ``bad'' yet have low \textit{specificity} (i.e. high \textit{oversensitivity}). Given classes of equal utility, we can have (a) perfect discrimination but poor specificity; (b) perfect specificity but poor discrimination; (c) perfect on both using \textit{discretization}.}
  \label{fig:variance_tradeoff}

\end{figure}

Benchmarks suggest that the  problem of reward modeling is nearly solved. Reward models achieve very high agreement with annotated preferences; on RewardBench 1 and 2 \citep{Lambert2024RewardBenchER, Malik2025RewardBench2A}, respectively, the top models achieve agreement of 94\% and 84\%. We argue that \textit{reward modeling is not solved}, and understanding this requires rethinking how we \textit{use} and \textit{evaluate} reward models.
RM evaluation typically assumes one response is always better than the others \citep{Lambert2024RewardBenchER,liu2024rm,Malik2025RewardBench2A}.\footnote{One important exception is the \textit{``Ties''} subset of RewardBench 2, which we discuss in greater length in \autoref{sec:intrinsic_eval}. This subset is a proxy for oversensitivity, but it only contributes 1/16th of the average score on Reward Bench.}  Similarly, many RL algorithms optimize the difference in each response's reward and the average among a batch of responses.

\begin{figure}[t]
    
  \centering
\includegraphics[trim={0.5cm 0.1cm 0.5cm 0.2cm}, clip, width=0.85\linewidth]{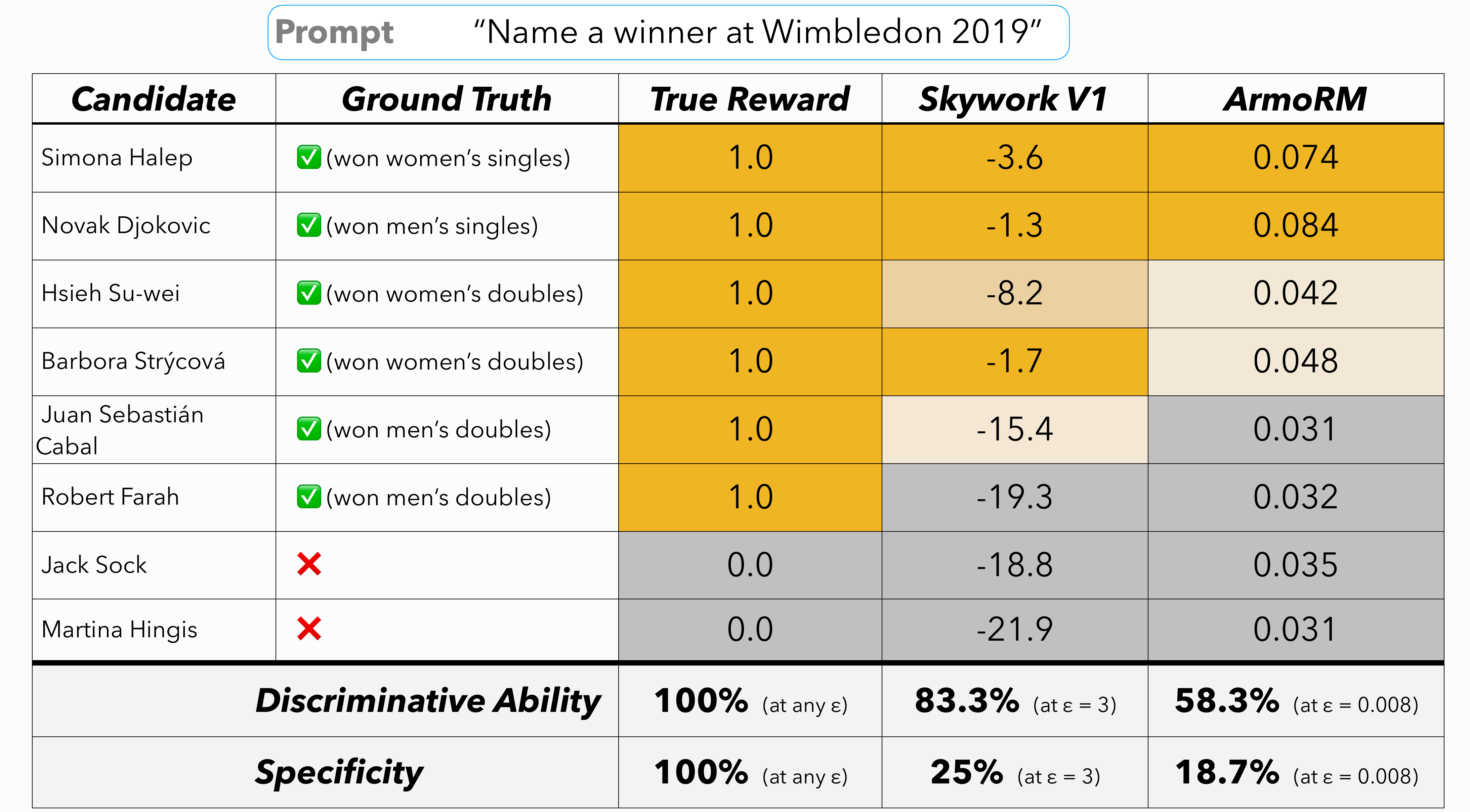}
      \caption{We propose measuring reward models using their \textit{discriminative ability} at distinguishing good responses from bad and their \textit{specificity} at identifying equally-good responses. Leading RM's \citep{Liu2024SkyworkRewardBO, ArmoRM} show strong discriminative ability but poor specificity. %
      }
  \label{fig:indeterminacy_example}
\vspace{-8pt}
\end{figure}

This assumption is clearly false. Most prompts can be answered in many equally-useful ways, such as ``Name a winner at Wimbledon 2019'' (shown in \autoref{fig:indeterminacy_example}). Within this group of equally-useful responses (e.g. \textit{the winner of women's singles} versus \textit{the winner of men's doubles}), average human preferences should usually be equal, and if they are not, it is an  artifact of ``rating indeterminacy'', where preferences arise from personal context, subjective interpretations, or harmful biases, rather than objective  differences \citep{guerdan2025validating}. Reward model training is also an imperfect process, and the resultant RM's trained on this data consequently models both objective utility signals and spurious rating artifacts like implicit biases \citep{Liu2024SkyworkRewardBO, liu2025skywork, ArmoRM, GRM}. 
When a reward model assigns different scores to equally good responses, it is not being \textit{discriminative} --- it is being \textit{oversensitive}. We show that oversensitivity is not merely noise to be averaged out, but provides a learnable signal that models can exploit.

Our work builds on recent observations that accurate reward models are not always good teachers for reinforcement learning~\citep{Chen2024TheAP}. One hypothesis offered in prior work is that in addition to \textit{accuracy}, RM's must show \textit{high variance over the reward space} to provide sufficient learning signal \citep{Razin2025WhatMA,Yang2025AcceleratingRT}. We suggest that what matters is not variance per se, but \textit{where} the variance comes from. Variance among responses of different utility is beneficial, while variance between responses of equal utility is harmful. \autoref{fig:variance_tradeoff} illustrates this distinction: blindly maximizing variance (center) sacrifices discriminative power and blindly maximizing accuracy (left) minimizes variance. The right panel shows that a careful discretization can  optimize both discriminative ability and \textit{specificity} (the complement of oversensitivity).  In \autoref{sec:discretization_minimizes_oversensitivity}, we prove this is possible under certain conditions. 

In a practical RL setting, we need to estimate these discretization thresholds without any prior knowledge.
We propose an algorithm to achieve this by estimating the predictive variance of the reward model using Monte Carlo dropout \citep{pmlr-v48-gal16, Gao2021SimCSESC} and using this to cluster responses into groups. We first evaluate our algorithm using the \textit{Ties} subset of RewardBench 2 \citep{Malik2025RewardBench2A}, where we find that, over a set of popular RM's, we consistently reduce the oversensitivity of the reward model at the modest expense of  discriminative ability, leading to an improvement in the average of the two. We then simulate an environment with a mixture of a primary reward (instruction following) and a secondary spurious (stylistic) reward. Optimizing this mixture directly leads to major stylistic overoptimization at the expense of the primary reward, but discretization allows models to maintain primary task efficacy. We lastly consider a realistic multi-task RL scenario of training a policy on unlabeled prompts. Comparing popular reward models with their discretized variants across IFEval, GSM8K, and MATH, discretization is never significantly worse than learning from the raw reward, and it is frequently much better. This suggests the potential of \textit{discretization via reward clustering} as a replacement for standard RL when using learned rewards.

\section{Problem Formulation}
\label{sec:problem}

\subsection{Accuracy, Discriminative Ability, and Oversensitivity}
\label{sec:definitions}

\paragraph{Definitions} We are trying to learn a policy $\pi: \mathcal{S} \to \mathcal{A}$.

For a prompt\footnote{We describe our setting as a single user prompt in single-turn interaction between user and model, but the state given to the model could similarly be a prompt prepended with full conversational context.} $x \in \mathcal{S}$ and response $y \in \mathcal{A}$, define\footnote{We will use the subscript notation $u_x(y)$ for brevity.} the ``true utility'' as an integer $u(x, y): \mathcal{S} \times \mathcal{A} \to [m_x] \subset \mathbb{Z}$.  $m_x$ is a random variable that depends on $x$, reflecting the number of equivalence classes of equally-good responses that exist for a given prompt $x$. Note that $u(x,y)$ is integer-valued; we explicitly consider the scenario where only a finite number of such equivalence classes exist. This means that utility must be countable (i.e. there is a limit to the differences between responses that humans can perceive) and bounded (i.e. there exist classes of  ``best possible'' and ``worst possible'' responses); we argue these assumptions are realistic \citep{guerdan2025validating, elangovan2025beyond}.

Our goal is to learn to produce responses that optimize our utility function:  ${\argmax}_y u(x, y)$. However, for a given $x$, the true $u$ is not known. Instead, we have access to a \textit{learned reward model}  $r: \mathcal{S} \times \mathcal{A} \to \mathbb{R}$.

In this paper, we introduce two constructs --- $\textit{discriminative ability}$ and $\textit{specificity}$ --- to measure reward model efficacy. The  \textit{discriminative ability} of $r$ is defined as $P\left(r_x(a) > r_x(b) | u_x(a) > u_x(b)\right)$. If this probability is 1 for a given reward model $r$, it will show perfect accuracy on reward model evaluation benchmarks like RewardBench 1 \citep{Lambert2024RewardBenchER}. We are also interested in \textit{specificity}, which is defined as $P\left(r_x(a) = r_x(b) | u_x(a) = u_x(b)\right)$. This is the complement of oversensitivity \citep{James2013AnIT}. Practically, we care about these quantities at a \textit{tolerance} $\epsilon \geq 0$: reward gap smaller than $\epsilon$ is assumed to be insignificant and treated as a tie. Under this tolerance, we redefine discriminative ability and specificity as $D_{r}(\epsilon) := P\left(r_x(a) > r_x(b) + \epsilon \mid u_x(a) > u_x(b)\right)$ and $\mathrm{Spec}_{r}(\epsilon) := 1 - P\left(|r_x(a) - r_x(b)| > \epsilon \mid u_x(a) = u_x(b)\right)$, respectively.

These constructs are more  granular than the prior notion of \emph{reward model accuracy}, formalized by  \citet{Razin2025WhatMA}) as  $\mathbb{E}_{a, b \sim \mathcal{A}} [\mathbbm{1}[\sign(r_x(a) - r_x(b)) = \sign(u_x(a) - u_x(b))]$. 

\begin{proposition}
\label{prop:first}
A reward's accuracy is the weighted sum of discriminative ability and specificity at $\epsilon = 0$.
\end{proposition}
\paragraph{Proof}
Accuracy is
$\mathbb{E}_{a, b} [\mathbbm{1}[\sign(r_x(a) - r_x(b))] = \sign(u_x(a) - u_x(b))]$.

We show in \autoref{app:accuracy_oversensitivity_calculation} that accuracy is equal to:
\begin{align*}
& P[r_x(a) = r_x(b) | u_x(a) = u_x(b)] \quad \times P\left[u_x(a) = u_x(b)\right]\tag*{\text{\emph{(specificity)}}}\\
& + P\left[r_x(a) > r_x(b) | u_x(a) > u_x(b)\right] P[u_x(a) > u_x(b)] \\
& + P\left[r_x(a) < r_x(b) | u_x(a) < u_x(b)\right] P[u_x(a) < u_x(b)] 
\tag*{\text{\emph{(discriminative ability)}}}\\ 
\end{align*}

In practice, RM evaluation focuses on discriminative ability. If utility is continuous, $P[u_x(a) = u_x(b)] = 0 \implies$ \textit{accuracy} = \textit{discriminative ability}. In \autoref{sec:discrim_and_oversensitivity}, we show conditions in which RM's with excellent discriminative ability can also be highly oversensitive. In \autoref{sec:discretization_minimizes_oversensitivity}, we will show that proper discretization can preserve discriminative ability while minimizing oversensitivity.

\setlength{\belowcaptionskip}{-20pt}
\begin{wrapfigure}[19]{r}{0.45\textwidth}
\vspace{-22pt}
  \centering
\includegraphics[width=0.99\linewidth]{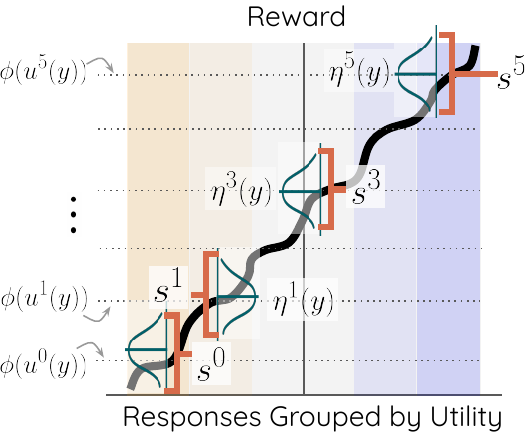}
  \caption{Discretization requires estimating regions of equivalent utility in reward space.  $\phi(u^i(a))$ marks the mean reward within class of responses with equal utility. $s^i$ is the spread of rewards corresponding to that utility class, given by $r(y) = \phi(u(y)) + \eta(y)$.}
  \label{fig:illustrating_discretization}
\end{wrapfigure}

To simplify our analysis, we assume our reward $r$ is a linear function of utility with a bounded amount of bias: $r_x(y) =$
$\phi_x(u_x(y)) + \eta_x(y)$,  where $\phi_x(v) = s_x v+d_x$ and $\eta_x: \mathcal{A} \to \mathbb{R}$ is \textit{reward noise} and $\phi_x$ is a linear function. Practically, we consider the case that $\eta_x(y)$ is easier to learn than $\phi_x(u_x(y))$, leading to \textit{reward hacking}.

We will assume that $\eta_x$ is bounded such that $|\eta_x| < s_x/2$, to ensure that this noisy reward model maintains perfect discriminative ability under our utility function. For ease of explanation, here we assume that $\eta_x(a) \sim \text{Unif}(-s_x/2, s_x/2)$. We show in \autoref{sec:gaussian_noise_relaxation} that the same conclusions hold after relaxing this assumption to generate $\eta$ from a Gaussian distribution.\footnote{We use reward noise $\eta$ as if it is a random variable, but it is actually a learnable, prompt-dependent function of a response. Consider that responses are sampled from a reference distribution $y \sim \bm{\pi}$. Then, within a utility class $c$, we assume $\eta_x(y) \sim \mathcal{N}(0, \sigma_x^2)$ given $u_x(y) = c$. Despite being a learnable function, $\eta$ allows a probabilistic interpretation over the distribution of responses.} Our practical algorithm in \S\ref{sec:discretization} supports that assumption. Intuitively, we assume that spurious or subjective preferences are present in reward models but their magnitude is much smaller than the differences in true utility. We illustrate the relationship between our reward model $r$, the utility function $u$, and the amount of noise $s$ in  \autoref{fig:illustrating_discretization}.

\subsection{Reward models with perfect discriminative ability can be highly oversensitive}
\label{sec:discrim_and_oversensitivity}

\begin{proposition}
    If $|\eta_x(a)| < s_x/2$, then $r$ has perfect discriminative ability at $\epsilon$ = 0.
\end{proposition}
\label{prop:perfect_discriminative_ability}

\paragraph{Proof}

\begin{align*}
&\text{ If } u_x(a) \neq u_x(b)\text{, then: } \sign(r_x(a) - r_x(b))\\
& = \sign(\phi(u_x(a)) + \eta_x(a) - \phi_x(u_x(b)) - \eta_x(b)) \\
& = \sign(s_x u_x(a) + d_x + \eta_x(a) - s_x u_x(b) - d_x - \eta_x(b))) \\
& = \sign(s_x (u_x(a) - u_x(b)) + \eta_x(a) - \eta_x(b)).
\end{align*}

Since $u$ is an integer-valued function, $u_x(a) \neq u_x(b) \implies |u_x(a) - u_x(b)| \geq 1$. Because $|\eta_x(a)| < s_x/2 \implies |\eta_x(a) - \eta_x(b)| < s_x$, then $\sign(r_x(a) - r_x(b))$ $= \sign(s_x\times(u_x(a) - u_x(b)) + \eta_x(a) - \eta_x(b)) > \sign(s_x\times(u_x(a) - u_x(b)))$.

This means $u_x(a) > u_x(b) \implies r_x(a) > r_x(b)$, and therefore $r$ has perfect discriminative ability.

\begin{proposition}
\label{prop:raw_discriminative_with_margin}
 If $\eta_x(a), \eta_x(b) \sim \text{Unif}(-s_x/2, s_x/2)$, then for any tolerance $\epsilon \in [0, 1]$ (measured in units of $s_x$ --- ``utility units''), the discriminative ability of the reward on \emph{adjacent} utility classes is
 \[
 D_{r}(\epsilon) = P(r_x(a) > r_x(b) + \epsilon | u_x(a) = u_x(b) + 1) = 1 - \frac{\epsilon^2}{2}.
\]

\end{proposition}

\paragraph{Proof}
$r_x(a) - r_x(b) = \eta_x(a) - \eta_x(b) + s_x$. Then
$(r_x(a) - r_x(b))/s_x \;=\; 1 + \eta_x(a) / s_x - \eta_x(b)/s_x$. Since $\eta_x(a)/s_x$, $\eta_x(b)/s_x$ $\sim$ $\text{Unif}(-1/2, 1/2)$, then $P(r_x(a) - r_x(b) > \epsilon s_x) = P((r_x(a) - r_x(b))/s_x > \epsilon) = P(\eta_x(a)/s_x - \eta_x(b)/s_x) > \epsilon  - 1) := P(x - y >  \epsilon - 1) $ where $x, y \sim \text{Unif}(0,1)$. By the CDF of the symmetric triangular distribution, $P(x - y <  \epsilon - 1) $ $=$ $((\epsilon - 1) + 1)^2/2$ = $\epsilon^2/2$. Then, $P(r_x(a) > r_x(b) + \epsilon s_x) = 1 - \epsilon^2/2$.

We assumed adjacent utility classes: $u_x(a) = u_x(b) + 1$. This is the worst-case scenario. When utility classes are further apart, $P(r_x(a) - r_x(b) > \epsilon s_x) \geq P\bigl(\eta_x(a)/s_x - \eta_x(b)/s_x > \epsilon  - 2\bigr) \geq P\bigl((\eta_x(a)- \eta_x(b))/s_x > -1\bigr) = 1$.

\begin{proposition}
\label{prop:raw_reward_oversensitivity_under_uniform}
 If $\eta_x(a), \eta_x(b) \sim \text{Unif}(-s_x/2, s_x/2)$, then for any tolerance $\epsilon \in [0, 1)$ (in units of $s_x$),
    \[
      \mathrm{Spec}_{r}(\epsilon) := 1 - P\bigl(|r_x(a) - r_x(b)| > \epsilon  \mid u_x(a) = u_x(b)\bigr) = \epsilon (2-\epsilon) <  1.
    \]
\end{proposition}

\paragraph{Proof}

\begin{align*}
& 1 - P\bigl(|r_x(a) - r_x(b)| > \epsilon \mid u_x(a) = u_x(b)\bigr)
   = 1 - P\bigl(|\eta_x(a) - \eta_x(b)| \geq \epsilon\bigr) \\
& \quad = 1 - 2P\bigl(\eta_x(a) - \eta_x(b) \geq \epsilon\bigr)
   = 1 - 2 \cdot \tfrac{(1-\epsilon)^2}{2} = 1 - (1-\epsilon)^2 = \epsilon (2 - \epsilon).
\end{align*}

Therefore, the raw reward model is oversensitive with probability $(1-\epsilon)^2$, which is strictly positive for every tolerance $\epsilon < s_x$, though this likelihood decreases quadratically as $\epsilon$ increases.

\subsection{There exists a discretization that maintains perfect discriminative ability}
\label{sec:discretization_can_maximize_discriminative_ability}
We will post-process the output of a reward model to \textit{discretize} the rewards. For simplicity, we'll assume here that our discretization is \textit{binary}, where we fix a single threshold $\tau_x$ for binarizing the reward, though our findings can be generalized to multi-level utility functions. Define the operation $\text{\textit{disc}}_x(v) $ as $\mathbbm{1}[v > \tau_x]$. Then:
\begin{align*}
&  D_{\textup{disc}}(\epsilon) := P\bigl(\text{disc}_x(r_x(a)) > \text{disc}_x(r_x(b)) + \epsilon \mid u_x(a) > u_x(b)\bigr)  = P\bigl(r_x(b) \leq \tau_x < r_x(a) \mid u_x(a) > u_x(b)\bigr) \\
& = P\bigl(\phi_x(u_x(a)) + \eta_x(a) > \tau_x \mid u_x(a) > u_x(b)\bigr) \hspace{3pt} \times \hspace{3pt}   P\bigl(\phi_x(u_x(b)) + \eta_x(b) \leq \tau_x \mid u_x(a) > u_x(b)\bigr).
\end{align*}

This derivation is based on the fact that discretized rewards are integer-valued, so a gap of 1 exceeds any margin $\epsilon < 1$. Since $u_x$ is integer-valued and $u_x(a) > u_x(b)$, we then have $u_x(a) \geq u_x(b) + 1$, which implies $\phi_x(u_x(a)) \geq \phi_x(u_x(b)) + s_x$.

Given that $|\eta_x| < s_x/2$, the ranges of possible rewards for responses $a$ and $b$ are non-overlapping:
\begin{align*}
& r_x(a) \hspace{10pt} =  \hspace{10pt}  \phi_x(u_x(y)) + \eta_x(y)  \hspace{10pt} \geq  \hspace{10pt} \phi_x(u_x(b)) + s_x/2 > r_x(b).
\end{align*}

Therefore, if the threshold satisfies $\tau_x \in [\phi_x(u_x(b)) + s_x/2, \phi_x(u_x(a)) - s_x/2]$, then $P(r_x(a) > \tau_x) = 1$ and $P(r_x(b) \leq \tau_x) = 1$, giving  \hspace{2pt} $D_{\textup{disc}}(\epsilon) =
P(\text{disc}_x(r_x(a)) > \text{disc}_x(r_x(b)) \mid u_x(a) > u_x(b)) = 1$.

\subsection{Discretization can minimize oversensitivity}
\label{sec:discretization_minimizes_oversensitivity}
\begin{proposition}
\label{prop:discretized_reward_oversensitivity_under_uniform}
    If $\eta_x(a) \sim \text{Unif}(-s_x/2, s_x/2)$, then oversensitivity $:= 1- \mathrm{Spec}_{\textup{disc}} = P\bigl(|\textup{disc}_x(r_x(a)) - \textup{disc}_x(r_x(b))| > \epsilon \mid u_x(a) = u_x(b)\bigr) = \max\!\bigl(0,\ \tfrac{1}{2} - 2\bigl((\tau_x - \phi_x(u_x(a)))/s_x\bigr)^2\bigr)$.
\end{proposition}

We give a proof of this proposition in \autoref{app:discretized_reward_oversensitivity_under_uniform_proof}.

\begin{theorem}
$\mathrm{Spec}_{\textup{disc}}(\epsilon) > \mathrm{Spec}_{\textup{raw}}(\epsilon)$ (proper discretization reduces oversensitivity)

\end{theorem}

\paragraph{Proof}

By Proposition \ref{prop:discretized_reward_oversensitivity_under_uniform}, $\mathrm{Spec}_{\textup{disc}}(\epsilon) =  1 - \max(0, \frac{1}{2} - 2((\tau_x - \phi_x(u_x(a))/s_x))^2) = \min(1, \frac{1}{2} + 2((\tau_x - \phi_x(u_x(a))/s_x))^2)$. By Proposition \ref{prop:raw_reward_oversensitivity_under_uniform}, $\mathrm{Spec}_{\textup{raw}} = \epsilon (2 - \epsilon) = 1 - (1 - \epsilon)^2$.

These quantities are not directly comparable, as $\mathrm{Spec}_{\textup{raw}}(\epsilon)$ depends on $\epsilon$ and, if we assume $\epsilon < s_x$, $\mathrm{Spec}_{\textup{disc}}(\epsilon)$ does not. However, raw reward models show imperfect specificity (i.e. nonzero oversensitivity):  $1 - \mathrm{Spec}_{\textup{raw}} = (1 - \epsilon)^2  > 0$. In contrast, there exist discretization thresholds with zero oversensitivity. If $(\tau_x - \phi_x(u_x(a)))/s_x \leq -\tfrac{1}{2}$ or $(\tau_x - \phi_x(u_x(a)))/s_x \geq \tfrac{1}{2}$, then 
$\mathrm{Spec}_{\textup{disc}}(\epsilon) =  1$.

In \autoref{sec:discretization_can_maximize_discriminative_ability}, we showed that if $(\tau_x - \phi_x(u_x(a)))/s_x \leq -\tfrac{1}{2}$ or $(\tau_x - \phi_x(u_x(a)))/s_x \geq \tfrac{1}{2}$, then perfect discriminative ability is maintained. Therefore, $\tau_x = \phi_x(u_x(a))- s_x/2$ or $\tau_x =  \phi_x(u_x(a)) + s_x/2$ optimizes both discriminative ability and specificity. In the case of a binary utility function, if we set $\tau_x = \phi_x(0) + \phi_x(1))/2$, then the discretized reward attains \textit{perfect discriminative ability} and \textit{perfect specificity}. In other words, an optimal discretization is achieved by setting thresholds that maximize the distance between each threshold and the mean reward of each adjacent equivalence class.

\subsection{Discretization maximizes the average of discriminative ability and specificity}
\label{sec:uniform_sum}

Because the margin $\epsilon$ now enters discriminative ability and specificity in the same way, we can summarize a reward model by a single combined score $T_{r}(\epsilon) \;:=\; (D_{r}(\epsilon) + \mathrm{Spec}_{r}(\epsilon))/2$, which attains its maximum of $1$ exactly when the reward has perfect discriminative ability and zero oversensitivity.

\begin{theorem}
\label{thm:uniform_sum}
Under a binary utility model, discretization can attain the maximal combined score at every $\epsilon$. If we fix the optimal midpoint threshold $\tau_x$, with $|\tau_x - \phi_x(u)|/s_x = 1/2$ for each adjacent class $u$, then for every $\epsilon \in [0,1)$, $T_{\textup{disc}}(\epsilon) \;=\; 1 \text{(maximum)}$, while the raw reward model will never exceed $T_{r}(\epsilon) > 5/6$.

Therefore, a proper discretization improves over the raw rewrd model at every tolerance: $T_{\textup{disc}}(\epsilon) - T_{r}(\epsilon) \ge \tfrac{1}{6}$ for all $\epsilon$, and the deficit approaches $\tfrac12$ as $\epsilon \to 0$.
\end{theorem}

\paragraph{Proof}
First consider the total score for the discretized reward. By \autoref{sec:discretization_can_maximize_discriminative_ability}, $D_{\textup{disc}}(\epsilon) = 1$ for $\epsilon \in [0,1)$. By Proposition \ref{prop:discretized_reward_oversensitivity_under_uniform} at $|\tau_x - \phi_x(u)|/s_x = 1/2$, the discretized oversensitivity is $\max(0, 1/2 - 2\cdot(1/2)^2) = 0$, so $\mathrm{Spec}_{\textup{disc}}(\epsilon) = 1$. Hence $T_{\textup{disc}}(\epsilon) = (1 + 1)/2 = 1$.

Now consider the total score for the raw reward. Proposition \ref{prop:raw_discriminative_with_margin}, $D_{r}(\epsilon) = 1 - (1/2)\epsilon^2$, and by Proposition \ref{prop:raw_reward_oversensitivity_under_uniform}, $\mathrm{Spec}_{r}(\epsilon) = \epsilon (2 - \epsilon)$, so $T_{r}(\epsilon) = -\tfrac34\epsilon^2 + \epsilon + \tfrac12$. $T'(\epsilon) = -\tfrac32\epsilon + 1$ and $T''(\epsilon) = -\frac32$. Since $T$ is concave down and its first derivative is zero at $\epsilon = \tfrac23$, its maximum value is is $T(2/3) = 1/2 + 2/3 - 1/3 = 5/6$. Thus $T_{r}(\epsilon) \le \tfrac56 < 1 = T_{\textup{disc}}(\epsilon)$. This gap is minimized at $\epsilon=\tfrac23$ with a value of $\frac16$ and rises to $\tfrac12$ as $\epsilon \to 0$.

If we relax our noise model to be Gaussian rather than bounded uniform noise, the condition that $T_{\textup{disc}} > T_{\textup{raw}}$ is maintained (see \autoref{thm:gaussian_net_benefit} in \autoref{app:gaussian_noise}).

\subsection{Related formulations}
Prior work has explored reward model pathologies that relate to oversensitivity.  \citet{InfoRM}, \citet{PBRR}, and others consider general \textit{reward misspecification}. \citet{InfoRM} train RMs with an information bottleneck to address this. \citet{PBRR} ``repair'' RM's for sequential decision-making in tabular environments. \citet{DPL} identify hidden context in multi-annotator preference data (a potential cause of oversensitivity) by learning a distribution rather than a point estimate reward. The interventions proposed in these prior works all training a new reward model. We believe that  on-the-fly discretization is superior due to \textit{flexibility} and \textit{generalization} in arbitrary scenarios involving a neural reward. Most similar to ours, \citet{afsharrad2026binarypreferencesprincipledframework} concurrently consider ordinal reward models (which are structurally identical to discretized reward models) --- however, they propose obtaining such a discretized RM via a new training procedure for reward models using ordinal preference magnitude labels, and their goal is to improve overall reward model accuracy (without defining or measuring oversensitivity separately from overall accuracy). Lastly, \citet{dark_side} consider \textit{false positive rewards} (in the setting of video game agents), which they detect with an external embedding-based judge model --- in contrast, our method provides a model-agnostic intervention requiring no external judge mechanism.

\section{Discretization via \textit{Reward Clustering}}
\label{sec:discretization}
We now describe an algorithm called \textit{reward clustering} that estimates the posterior distribution of each reward to group responses likely to be equally useful, illustrated by \autoref{fig:illustrating_discretization}.

\paragraph{Discretization as Clustering} We treat reward discretization as a 1-D clustering problem. Given a set of rewards $r_1, \ldots, r_n$ for a batch of responses $a_1, \ldots, a_n$, we estimate $P(\lvert r_i - r_j \rvert < \Delta)$ for each pair $r_i, r_j$. We then perform hierarchical clustering using complete linkage over these pairwise distances and cut the resulting dendrogram such that, for all $i,j$ in each final cluster, $P(\lvert r_i-r_j \rvert < \Delta) > p^*$, where $\Delta$ and $p^*$ are hyperparameters. The responses in each cluster are then assigned sequential integer rewards corresponding to the ordinal rank of their cluster's mean reward.

\paragraph{Estimating Equivalent Rewards via MC Dropout}
In \S\ref{sec:discrim_and_oversensitivity}, we assumed uniform noise to provide hard guarantees. In practice, we consider a more realistic setting where rewards are approximately Gaussian (which is often the case with Bradley-Terry RM's   \citet{Sun2024RethinkingBM}). As we show in \autoref{app:gaussian_noise}, the core insight --- discretization reduces oversensitivity in the presence of responses with equal utility --- holds under both distributions. Assuming the reward estimates are independent Gaussians,
$r_i \sim \mathcal{N}(\hat{\mu}_i, \hat{\sigma}_i^2)$, the difference in rewards is distributed as 
$r_i - r_j \sim \mathcal{N}\left(\hat{\mu}_i - \hat{\mu}_j, \hat{\sigma}_i^2 + \hat{\sigma}_j^2\right)$.

Then, the probability that two rewards are within $\Delta$ of each other is the probability that their difference falls in $(-\Delta, \Delta)$:
\begin{align*}
& P(|r_i - r_j| < \Delta) = P(-\Delta < r_i - r_j < \Delta) = \Gamma\left(\frac{\Delta - (\hat{\mu}_i - \hat{\mu}_j)}{\sqrt{\hat{\sigma}_i^2 + \hat{\sigma}_j^2}}\right) - \Gamma\left(\frac{-\Delta - (\hat{\mu}_i - \hat{\mu}_j)}{\sqrt{\hat{\sigma}_i^2 + \hat{\sigma}_j^2}}\right)
\end{align*}
where $\Gamma$ is the cumulative distribution function of the standard normal distribution.

We can assume that a reward sampled from our reward model is a reasonable estimate of $\hat{\mu}$, but we need to estimate the predictive variance $\hat{\sigma}^2$. 
We look to Monte Carlo (MC) Dropout \citep{pmlr-v48-gal16, Gao2021SimCSESC} as a solution. For each response $a_i$, we perform $T$ stochastic forward passes through the reward model with a dropout probability of \textit{d}, yielding reward samples $\{r_i^{(1)}, \ldots, r_i^{(T)}\}$. Increasing the value of $T$ ought to improve the estimate of variance, but this linearly increases the cost of reward computation. We can then approximate\footnote{\citet{pmlr-v48-gal16} show that dropout training approximates variational inference over network weights, and thus sampling with dropout at test time gives an estimate of epistemic uncertainty. Reward models are usually trained without dropout; in this case MC dropout is not an exact estimate of epistemic uncertainty, but we find it is a useful estimate in practice.
} predictive variance: $\hat{\sigma}_i = \sqrt{\frac{1}{T-1}\sum_{t=1}^T (r_i^{(t)} - \hat{\mu}_i)^2}$.

\paragraph{Practical Implementation} We implement reward clustering using the OpenRLHF library \citep{openrlhf}. Reward clustering has 4 hyperparameters: $\Delta$ (the difference needed to consider two reward values equivalent), $p^*$ (the minimum likelihood of reward equivalence needed to merge a given response into a cluster), $d$ (dropout probability), and  $T$ (number of samples used for dropout). In all our experiments, we choose $d=0.02$ (which provides sufficiently diverse samples) and $T=4$. We show in \autoref{sec:num_dropout_samples} that performance surprisingly does not increase with the number of samples taken via MC dropout. When training on nodes with 8 x H100 GPUs with Ray \citep{ray}, discretization increases average GRPO runtime by 15\% (training throughput over 6 runs drops from $64.3 \pm 8.0$ prompts per minute to $55.8 \pm 7.9$).

\section{Experiments}
\label{sec:experiments}

\subsection{Reward models exhibit oversensitivity}
\label{sec:intrinsic_eval}

We first explore how to modulate the discriminative ability and oversensitivity of reward models using the \textit{``Ties''} subset of \textit{RewardBench 2} \citep{Malik2025RewardBench2A}. We compare ways to use the output of four leading reward models --- Skywork V1 \citep{Liu2024SkyworkRewardBO}, Skywork V2 \citep{liu2025skywork}, GRM \citep{GRM}, and ArmoRM \citep{ArmoRM}.

\paragraph{Metrics} RewardBench 2 \textit{``Ties''} consists of prompts with one or more \textit{chosen} response and three or more \textit{rejected} responses. 
The primary metric used is the weighted average of two \textit{accuracy}  and \textit{margin}, where \textit{accuracy} measures whether or not all ``chosen'' responses receive higher reward than any ``rejected'' response and \textit{margin} measures whether the difference between the worst-scoring ``chosen'' response and the best-scoring ``rejected'' response is greater than the difference between the best-scoring and worst-scoring ``chosen'' responses. The final metric is $0.6 \times \textit{accuracy} + 0.4 \times \textit{margin}$.

This metric permits substantial oversensitivity as long as it does not exceed a relative tolerance (``the margin''). Whether this tolerance is appropriate depends on how reward signals are consumed during RL. In practice, algorithms like GRPO, REINFORCE, and PPO optimize the normalized difference between a given reward and a ``baseline'' \citep{deepseekmath, Williams2004SimpleSG, Schulman2017ProximalPO}. This difference can provide a learnable signal regardless of how the baseline is computed.

To overcome these issues, we apply pairwise notions of \textit{discriminative ability} and \textit{specificity} (complement of oversensitivity), as defined in \S\ref{sec:definitions}, to this benchmark, by assuming that all ``chosen'' and all ``rejected'' responses have equal utility. To make our rewards scale-invariant, we normalize rewards at a per-prompt level over all responses for that prompt before computing our metrics. Discriminative ability and specificity both require a threshold $\hat{\epsilon}$ to determine tolerance for equivalence in the continuous reward space. We set $\hat{\epsilon} = 0.10$, which conceptually defines equivalent rewards as being within 10\% of the within-batch spread, as this value resulted in the strongest baseline using the raw reward model in most settings, and use this $\hat{\epsilon}$ when evaluating all post-processing methods.

\begin{table}[t]
\centering
\small
\setlength{\tabcolsep}{4pt}
\begin{tabular}{ l | c c c | c c c}
\toprule
Method & \multicolumn{3}{c|}{\textit{Proposed} Metrics} & \multicolumn{3}{c}{\textit{Standard} Metrics} \\
& Avg. & Discrim. & Specif. & Avg. & Acc. & Margin \\
\midrule
\multicolumn{7}{c}{\textit{Skywork V1}} \\
\midrule
Raw & 70.8  {\scriptsize ± 0.7}  & 99.2  {\scriptsize ± 0.2}  & 42.4  {\scriptsize ± 1.5}  & 80.2  {\scriptsize ± 2.5}  & \textbf{99.0}  {\scriptsize ± 0.8}  & 52.0  {\scriptsize ± 4.9} \\
Clipping & 71.6  {\scriptsize ± 0.8}  & 99.0  {\scriptsize ± 0.3}  & 44.2  {\scriptsize ± 1.6}  & \textbf{80.4}  {\scriptsize ± 2.7}  & 96.1  {\scriptsize ± 1.4}  & 56.9  {\scriptsize ± 5.0} \\
Ensemble & 70.7  {\scriptsize ± 0.7}  & \textbf{99.2}  {\scriptsize ± 0.2}  & 42.2  {\scriptsize ± 1.5}  & 78.2  {\scriptsize ± 2.7}  & 97.1  {\scriptsize ± 1.3}  & 50.0  {\scriptsize ± 5.0} \\
Binary & 64.0  {\scriptsize ± 0.2}  & 67.6  {\scriptsize ± 1.5}  & 60.4  {\scriptsize ± 1.7}  & 36.3  {\scriptsize ± 2.0}  & 2.9   {\scriptsize ± 1.2}  & \textbf{86.3}  {\scriptsize ± 3.6} \\
$\bigstar$ Clustered & \textbf{74.9}  {\scriptsize ± 0.8}  & 97.4  {\scriptsize ± 0.6}  & \textbf{52.5}  {\scriptsize ± 1.8}  & 69.8  {\scriptsize ± 3.5}  & 74.5  {\scriptsize ± 3.2}  & 62.8  {\scriptsize ± 5.2} \\
\midrule
\multicolumn{7}{c}{\textit{Skywork V2}} \\
\midrule
Raw & 71.7  {\scriptsize ± 0.8}  & 98.1  {\scriptsize ± 0.5}  & 45.2  {\scriptsize ± 1.6}  & 71.4  {\scriptsize ± 3.0}  & 88.2  {\scriptsize ± 2.5}  & 46.1  {\scriptsize ± 4.8} \\
Clipping & 70.7  {\scriptsize ± 0.8}  & 98.0  {\scriptsize ± 0.4}  & 43.4  {\scriptsize ± 1.6}  & \textbf{73.3}  {\scriptsize ± 2.9}  & 88.2  {\scriptsize ± 2.2}  & \textbf{51.0}  {\scriptsize ± 4.9} \\
Ensemble & 71.1  {\scriptsize ± 0.8}  & \textbf{98.1}  {\scriptsize ± 0.4}  & 44.2  {\scriptsize ± 1.6}  & 71.2  {\scriptsize ± 2.8}  & \textbf{89.2}  {\scriptsize ± 2.2}  & 44.1  {\scriptsize ± 4.8} \\
Binary & 63.7  {\scriptsize ± 0.2}  & 67.3  {\scriptsize ± 1.4}  & 60.1  {\scriptsize ± 1.6}  & 2.9   {\scriptsize ± 1.2}  & 2.9   {\scriptsize ± 1.2}  & 2.9   {\scriptsize ± 1.2} \\
$\bigstar$ Clustered & \textbf{73.2}  {\scriptsize ± 0.9}  & 96.0  {\scriptsize ± 0.8}  & \textbf{50.4}  {\scriptsize ± 1.8}  & 56.7  {\scriptsize ± 3.7}  & 67.7  {\scriptsize ± 3.3}  & 40.2  {\scriptsize ± 5.0} \\
\midrule
\multicolumn{7}{c}{\textit{GRM}} \\
\midrule
Raw & 69.2  {\scriptsize ± 0.7}  & \textbf{97.0}  {\scriptsize ± 0.7}  & 41.4  {\scriptsize ± 1.4}  & \textbf{67.7}  {\scriptsize ± 3.2}  & \textbf{83.3}  {\scriptsize ± 3.0}  & 44.1  {\scriptsize ± 5.0} \\
Clipping & 63.4  {\scriptsize ± 1.1}  & 92.9  {\scriptsize ± 1.6}  & 33.9  {\scriptsize ± 1.5}  & 44.9  {\scriptsize ± 3.0}  & 63.7  {\scriptsize ± 3.4}  & 16.7  {\scriptsize ± 3.7} \\
Ensemble & 63.3  {\scriptsize ± 0.7}  & 94.7  {\scriptsize ± 1.2}  & 32.0  {\scriptsize ± 1.0}  & 48.8  {\scriptsize ± 2.8}  & 71.6  {\scriptsize ± 3.0}  & 14.7  {\scriptsize ± 3.6} \\
Binary & 62.9  {\scriptsize ± 0.2}  & 66.5  {\scriptsize ± 1.4}  & 59.4  {\scriptsize ± 1.6}  & 33.1  {\scriptsize ± 2.0}  & 2.9   {\scriptsize ± 1.2}  & \textbf{78.4}  {\scriptsize ± 4.3} \\
$\bigstar$ Clustered & \textbf{80.6}  {\scriptsize ± 1.0}  & 82.6  {\scriptsize ± 2.1}  & \textbf{78.5}  {\scriptsize ± 1.9}  & 45.1  {\scriptsize ± 3.9}  & 35.3  {\scriptsize ± 3.8}  & 59.8  {\scriptsize ± 5.3} \\
\midrule
\multicolumn{7}{c}{\textit{ArmoRM}} \\
\midrule
Raw & 64.4  {\scriptsize ± 0.8}  & 93.3  {\scriptsize ± 1.3}  & 35.4  {\scriptsize ± 1.5}  & 49.4  {\scriptsize ± 2.9}  & 68.6  {\scriptsize ± 3.2}  & 20.6  {\scriptsize ± 3.8} \\
Clipping & 63.0  {\scriptsize ± 0.8}  & 93.0  {\scriptsize ± 1.2}  & 33.1  {\scriptsize ± 1.5}  & 48.6  {\scriptsize ± 3.1}  & 62.8  {\scriptsize ± 3.4}  & 27.5  {\scriptsize ± 4.1} \\
Ensemble & 64.2  {\scriptsize ± 0.8}  & \textbf{93.6}  {\scriptsize ± 1.3}  & 34.8  {\scriptsize ± 1.4}  & \textbf{49.6}  {\scriptsize ± 2.8}  & \textbf{69.6}  {\scriptsize ± 3.1}  & 19.6  {\scriptsize ± 3.8} \\
Binary & 62.2  {\scriptsize ± 0.2}  & 65.7  {\scriptsize ± 1.3}  & 58.7  {\scriptsize ± 1.5}  & 29.2  {\scriptsize ± 2.3}  & 2.9   {\scriptsize ± 1.2}  & \textbf{68.6}  {\scriptsize ± 5.1} \\
$\bigstar$ Clustered & \textbf{70.7}  {\scriptsize ± 1.2}  & 85.4  {\scriptsize ± 2.0}  & \textbf{56.0}  {\scriptsize ± 1.9}  & 43.7  {\scriptsize ± 3.2}  & 36.3  {\scriptsize ± 3.5}  & 54.9  {\scriptsize ± 5.1} \\
\bottomrule
\end{tabular}
\vspace{5pt}
\caption{Under our proposed new metrics on RewardBench 2,  \textit{reward clustering} improves the average of specificity and discriminative ability for every reward model. We use an equivalence tolerance of $\hat{\epsilon}=0.10$ in the normalized reward space for computing these metrics. On RewardBench 2's standard metrics, \textit{reward clustering} increases the margin between chosen and rejected responses for 3 out of 4 models. The best method on each metric in each setting is bolded. While we display standard deviations of each metric, we do not explicitly indicate significance for each comparison.}
\label{tab:rewardbench2_methods}
\vspace{8pt}
\end{table}

\paragraph{Hyperparameters} 
For reward clustering, we select values for the hyperparameters $\Delta$ and $p^*$ via a small (n=13), handwritten validation set of prompts labeled with equivalence classes of responses. We use $\Delta=10$ and $p^*=0.8$ for Skywork V1 and V2, $\Delta=5$ and $p^*=0.6$ for GRM, and $\Delta=0.08$ and $p^*=0.8$  for ArmoRM. 

\paragraph{Baselines} We compare other reward processing techniques with  \textit{reward clustering}. As baselines, we use \textit{raw} (using the RM directly) and \textit{clipping} (clipping the 20\% of tail rewards), along with \textit{ensembling} (taking the average of 4 rewards sampled via dropout) \citep{herding}, and \textit{binary thresholding} (using the median as a threshold for a simple model-agnostic discretization).

In \autoref{tab:rewardbench2_methods}, we see a tension between different evaluation paradigms. Reward clustering uniformly improves the mean of our proposed metrics, specificity and discriminative ability; this supports the theoretical claims made in \autoref{thm:uniform_sum} (in a simplified setting) and \autoref{thm:gaussian_net_benefit} (in the more practical setting of Gaussian-distributed rewards in each utility class). For every reward model, we can improve specificity in exchange for modest reductions in discriminative ability. On the standard metrics, reward clustering increases the margin metric relative over using the reward model directly in all settings except Skywork V2\footnote{Later on in \S\ref{sec:real_eval}, we will find Skywork V2 is also the RM on which we see the fewest gains in RL training experiments.}, but sometimes at the expense of  \textit{accuracy}. These default metrics do not fully capture these gains because they tolerate oversensitivity below some amount. Among baselines, we note that \textit{ensembling} does not improve over \textit{raw} --- the use of MC dropout in reward estimation does not intrinsically help. As performance of reward models on intrinsic capability benchmarks like RewardBench may be poorly correlated with their efficacy as teachers for RLHF \citep{liu2024rm, Frick2024HowTE, Malik2025RewardBench2A}, 
the downstream RL results in \S\ref{sec:synth_eval} and \S\ref{sec:real_eval} offer a more direct test: \textit{we find that gains on our proposed metrics translate to better policies}.

\subsection{Reward clustering inhibits overfitting to spurious effects}
\label{sec:synth_eval}

Can discretizing reward models help to distinguish \textit{signal} from \textit{noise}?  Modern reward models capture mixtures of useful preferences (task completion, readability) and harmful ones (sycophancy, \textit{goblin} references\footnote{\url{https://openai.com/index/where-the-goblins-came-from/}}, etc).  We simulate this scenario by specifying a primary goal (following precise instructions) and a secondary goal (to use non-committal, hedged language). This  is concretely defined as learning to produce ``hedging words''  (e.g. ``possibly'', ``maybe'') and avoid  ``intensifiers'' (e.g. ``very'', ``absolutely''). 

We operationalize this scenario with a preference dataset. This dataset is built from \texttt{allenai/RLVR-IFeval}. To support our primary goal, we sample responses from \textit{Llama-3.1-8B-Instruct} and group them into the following categories: \textit{(A)} solve the task using confident language; \textit{(B)} fail the task using confident language; \textit{(C)} solve the task using hedged language; \textit{(D)} fail the task using hedged language. For our primary goal, we construct response pairs where one response is verifiably correct and the other is verifiably incorrect but neither contains hedging nor intensifiers (A/B or C/D). For our secondary goal, we select equally-correct response pairs where one response is confident and the other is hedged  (A/C or B/D). We consider 90\%-10\% and 80\%-20\% mixtures of the ``Primary'' and ``Secondary'' subsets.\footnote{These mixtures would correspond to differences in the range of the noise component $\eta$ described in \autoref{sec:definitions}.} By design, the dataset is dominated by examples of the primary goal, but the secondary goal may be easier to learn and exploit. We then train Bradley-Terry reward models based on \textit{Llama-3.1-8B-Instruct} \citep{Liu2024SkyworkRewardBO} on both datasets. The resultant RM's encode both task correctness and linguistic hedging but in varying proportions.

\begin{figure}[t]
    
  \centering
\includegraphics[trim={0.2cm 0.2cm 0.0cm 2.5cm}, clip, width=\linewidth]{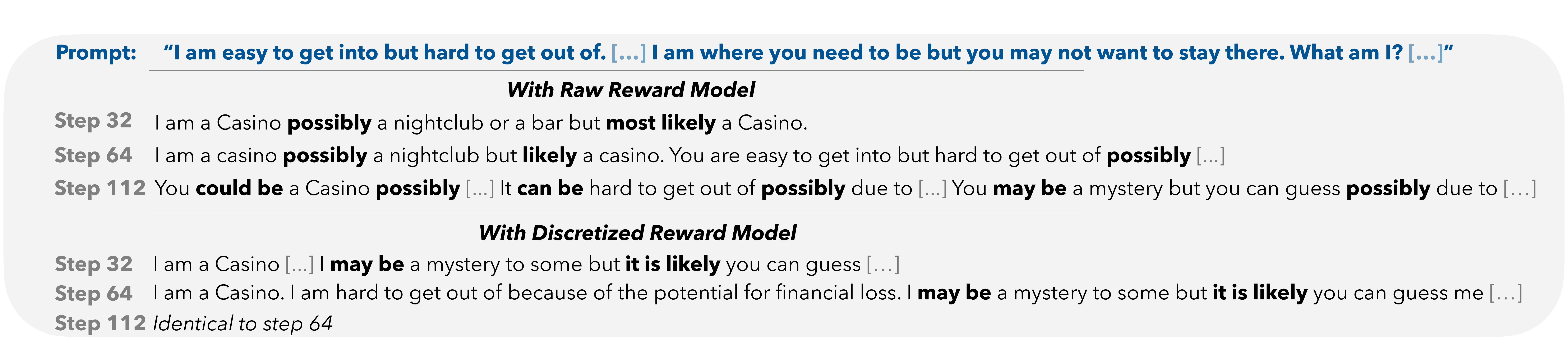}
  \caption{ On a real prompt from IFEval, we see that training directly on a reward containing both correct instruction-following rewards and spurious rewards (for ``linguistic hedging'') leads the model to excessively optimize for  hedged. In contrast, training on a discretized reward model leads to less hedging, and the model converges to a sensible response.
  }
  \label{fig:reward_hacking_example}
\vspace{10pt}
\end{figure}

\begin{figure}[t]
  \centering
  \begin{minipage}{0.49\linewidth}
    \includegraphics[width=\linewidth]{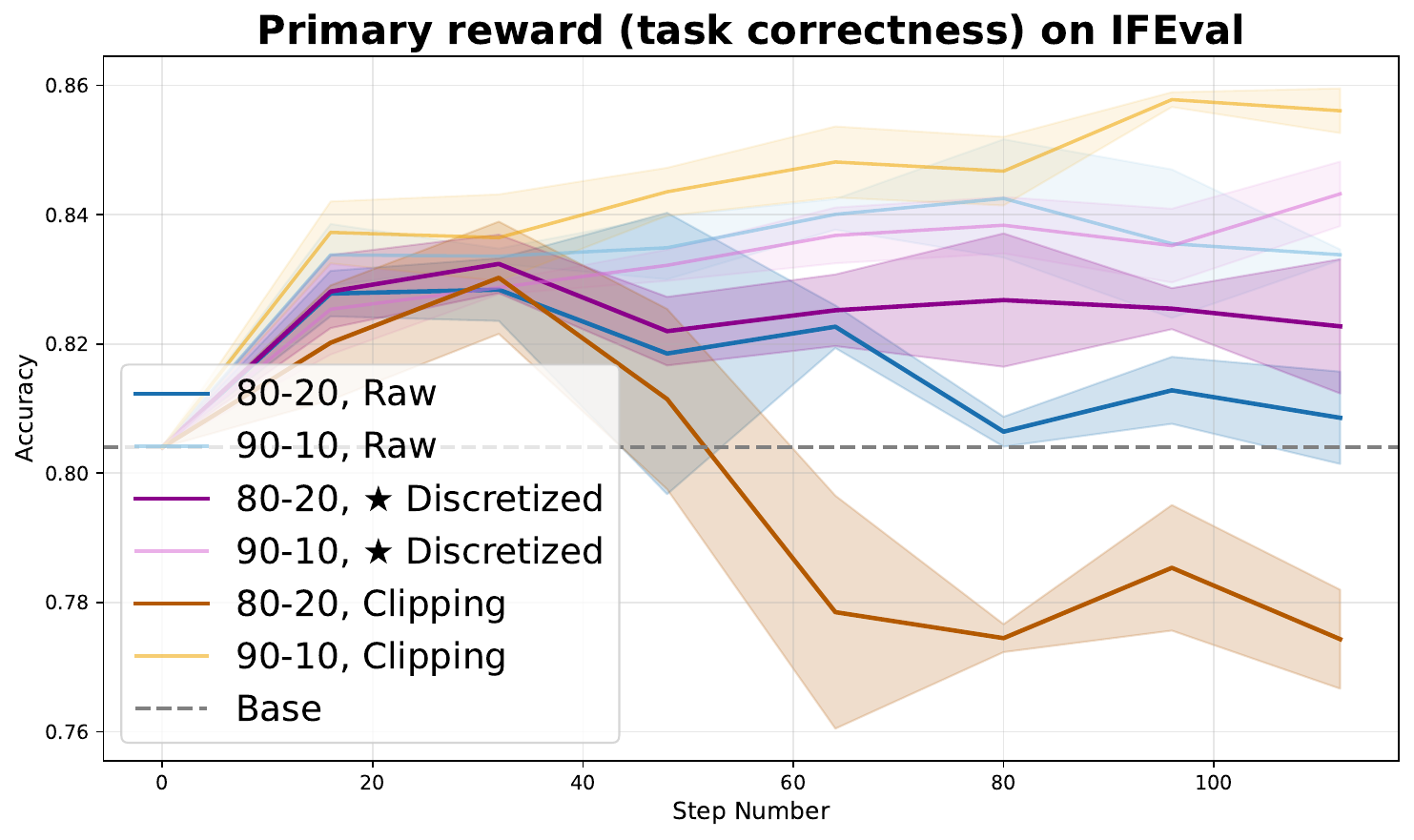}
  \end{minipage}
  \hfill
  \begin{minipage}{0.49\linewidth}
    \includegraphics[width=\linewidth]{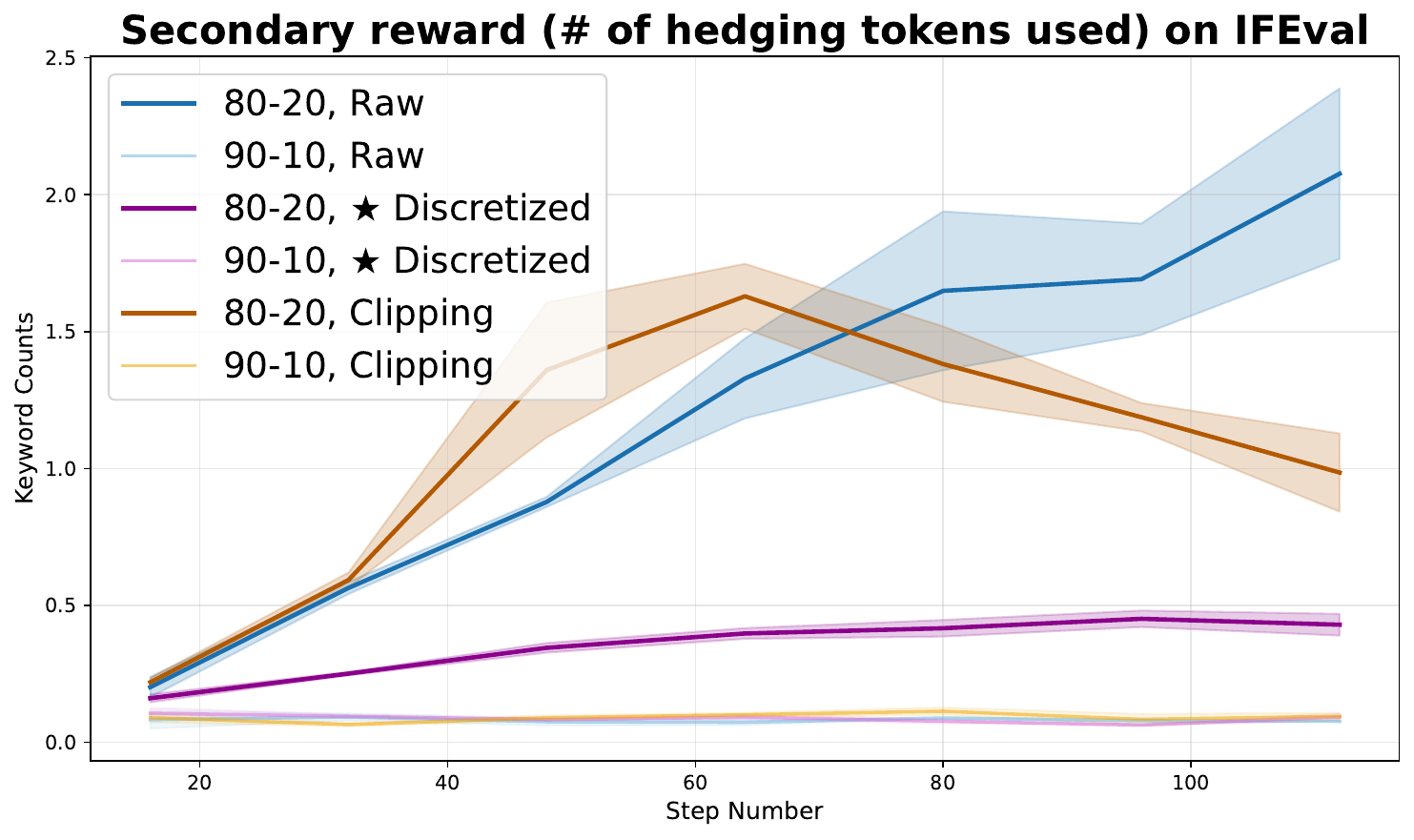}
  \end{minipage}
\caption{(Left) Policies trained via RL against mixed-effect reward models initially optimize task
correctness well, but performance degrades with continued training. Means and standard deviations
are reported over three runs. (Right) Policies trained on an 80-20 mixture learn to heavily exploit
the secondary reward (increasing ``hedging word'' usage); discretization curbs this overoptimization.
Clipping performs well in the 90-10 setting where the reward is better-specified but proves disastrous in the 80-20
setting, suggesting that when the hackable secondary reward carries greater weight, clipping may
cannibalize the primary reward rather than containing the exploit.}
\label{fig:perturbed_rlvr}
\vspace{15pt}
\end{figure}

In this environment, discretization enables
modest optimization of the secondary reward while safely preserving primary task performance. \autoref{fig:reward_hacking_example} illustrates how  a model trained on the 80\%-20\% mixture of instruction-following-based rewards and spurious rewards learns to  complete the task but hedges excessively. In \autoref{fig:perturbed_rlvr}, we see that, with a 20\% spurious preference ratio,
this overexploitation of the hedging reward comes at the expense of task
correctness; even at 10\%, optimizing the secondary reward still degrades performance (though without 
overt overoptimization). Clipping shows well in the 90-10 setting but proves disastrous
at 80-20, suggesting that a heavier hackable reward causes clipping to sacrifice the primary
objective.

\subsection{Reward clustering improves learning from real reward models}
\label{sec:real_eval}

We train Llama-3.1-8B-Instruct \citep{llama3} in a verifier-free  multi-task setup on unlabeled prompts. We use 30K combined prompts from the \texttt{allenai/RLVR-IFeval}, \texttt{allenai/RLVR-MATH}, and \texttt{allenai/RLVR-GSM} datasets and 30K prompts randomly sampled from WildChat \citep{Zhao2024WildChat1C}, which is truly non-verifiable. We use the same RM's and hyperparameters described in \S\ref{sec:intrinsic_eval}.
We train with two values of KL penalty with 3 random seeds each using GRPO with 8 rollouts \citep{deepseekmath} via OpenRLHF \citep{openrlhf}.
We evaluate on three in-distribution datasets, IFEval \citep{zhou2023instruction}, MATH \citep{hendrycksmath}, and GSM8K \citep{gsm8k}.

\begin{table}[t]
\centering
\small
\setlength{\tabcolsep}{3pt}
\resizebox{\linewidth}{!}{%
\begin{tabular}{l | cc | cc | cc | cc || cc}
\toprule
 Dataset & \multicolumn{2}{c|}{\textit{Skywork V1}} & \multicolumn{2}{c|}{\textit{Skywork V2}} & \multicolumn{2}{c|}{\textit{GRM}} & \multicolumn{2}{c||}{\textit{ArmoRM}} & \multicolumn{2}{c}{\textit{Aggregate}} \\
& Raw & Disc. & Raw & Disc. & Raw & Disc. & Raw & Disc. & Raw & Disc. \\
\midrule
\multicolumn{11}{c}{low KL penalty ($\beta = 0.01$)} \\
\midrule
GSM8K
    & 77.0 {\scriptsize ± 3.7} & \textbf{84.4} {\scriptsize ± 0.6}
    & 56.6 {\scriptsize ± 47.7} & \textbf{83.3} {\scriptsize ± 1.2}
    & 68.5 {\scriptsize ± 4.4} & \textbf{76.2} {\scriptsize ± 2.9}
    & 3.6 {\scriptsize ± 2.1} & 2.2 {\scriptsize ± 0.8}
    & 50.7 {\scriptsize ± 35.7} & 61.2 {\scriptsize ± 32.7} \\
  MATH
    & 48.0 {\scriptsize ± 1.3} & 49.6 {\scriptsize ± 0.5}
    & 50.1 {\scriptsize ± 1.6} & 47.8 {\scriptsize ± 1.5}
    & 39.5 {\scriptsize ± 0.9} & \textbf{45.9} {\scriptsize ± 1.4}
    & 19.1 {\scriptsize ± 27.0} & 38.0 {\scriptsize ± 2.3}
    & 34.4 {\scriptsize ± 20.3} & 34.8 {\scriptsize ± 21.0} \\
  IFEval
    & 52.6 {\scriptsize ± 3.0} & 55.8 {\scriptsize ± 1.0}
    & 77.9 {\scriptsize ± 1.7} & 79.1 {\scriptsize ± 1.4}
    & 43.5 {\scriptsize ± 0.8} & \textbf{63.0} {\scriptsize ± 1.9}
    & 53.0 {\scriptsize ± 1.9} & \textbf{77.8} {\scriptsize ± 0.8}
    & 46.7 {\scriptsize ± 23.4} & 63.1 {\scriptsize ± 13.3} \\
\midrule
\multicolumn{11}{c}{high KL penalty ($\beta = 0.05$ )} \\
\midrule
  GSM8K
    & 86.6 {\scriptsize ± 0.8} & 86.2 {\scriptsize ± 0.5}
    & 84.3 {\scriptsize ± 0.5} & 81.9 {\scriptsize ± 2.8}
    & 80.2 {\scriptsize ± 1.4} & 80.1 {\scriptsize ± 2.7}
    & 15.5 {\scriptsize ± 22.6} & 32.2 {\scriptsize ± 40.3}
    & 62.9 {\scriptsize ± 36.1} & 79.2 {\scriptsize ± 9.9} \\
  MATH
    & 51.9 {\scriptsize ± 0.3} & \textbf{52.6} {\scriptsize ± 0.4}
    & 51.6 {\scriptsize ± 0.8} & 33.4 {\scriptsize ± 28.3}
    & 48.5 {\scriptsize ± 0.3} & 49.0 {\scriptsize ± 0.5}
    & 20.8 {\scriptsize ± 16.8} & \textbf{45.1} {\scriptsize ± 1.5}
    & 39.5 {\scriptsize ± 19.8} & 42.3 {\scriptsize ± 15.1} \\
  IFEval
    & 76.3 {\scriptsize ± 0.6} & 77.0 {\scriptsize ± 1.3}
    & 81.8 {\scriptsize ± 0.3} & 82.5 {\scriptsize ± 0.5}
    & 66.1 {\scriptsize ± 0.5} & \textbf{69.1} {\scriptsize ± 0.7}
    & 60.0 {\scriptsize ± 3.3} & \textbf{80.0} {\scriptsize ± 1.0}
    & 64.5 {\scriptsize ± 21.3} & 75.1 {\scriptsize ± 7.3} \\
\bottomrule
\end{tabular}}
\caption{Discretization always significantly improves (10/24 comparisons greater than one standard deviation) or maintains efficacy (14/24 comparisons), with \textit{zero} regressions (up to significance). Our base model, \textit{Llama-3.1-8B-Instruct}, achieves 77.3 on GSM8K, 47.2 on MATH, and 77.5 on IFEval.}
\label{tab:downstream_performance}
\vspace{10pt}
\end{table}

\paragraph{Results} \autoref{tab:downstream_performance} shows discretization is never worse than the raw baseline (up to overlapping confidence intervals) and frequently much better. With two exceptions\footnote{The exceptions are ArmoRM with $\beta=0.01$ on MATH and one run of Skywork V2 on MATH with $\beta=0.05$. Both are a failure mode affecting our baseline (answering in-context examples rather than the target task) --- not unique to discretization.}, discretized rewards rarely leads to degenerate policies, unlike training on raw reward models. As a whole, these results suggest that \textit{discretized reward models} are suitable drop-in replacements for using reward models directly.

\section{Limitations}
\label{sec:limitations}
Our paper has three main limitations. While our algorithm makes few assumptions on model type, we only experiment with language models. Moreover, our theory assumes (in \S\ref{sec:problem}) that our reward model is a \textit{linear function of a binary utility function} plus \textit{noise that is either bounded uniform or Gaussian}. The linearity assumption may be sensible assumption (given a supposed utility function), but real-world utility functions often feature multiple equivalence classes of responses.  While our derivations should generalize cleanly to this setting, we have not shown so in this paper. In both cases, we also assume the reward noise has constant variance across utility classes, which simplifies our calculations but likely would not hold in practice. Lastly, all our experimental results are shown on a single base model, \textit{Llama-3.1-8B-Instruct}, and all RL training experiments use a single learning algorithm (GRPO; \citet{deepseekmath}).

\section{Conclusion}
\label{sec:conclusion}
Motivated by concerns that reinforcement learning supervised by reward models may lead to policies that learn undesirable behaviors, we introduce \textit{reward model oversensitivity} and find many popular RM's are highly oversensitive despite their strong discriminative ability. We show that, in theory, continuous-valued rewards with perfect discriminative ability must exhibit oversensitivity, but certain discretizations can mitigate this. We then describe \textit{reward clustering} which improves the tradeoff between specificity and discriminative ability and leads to better policies. Discretization raises the risk of reducing models' peak ability by slowing learning. However, our results suggest that discretization can retain the upside of learning from reward models while limiting  potential negative impacts.

\section{Acknowledgements}
We are grateful to Madian Khabsa for his role in scoping and formulating our problem. We thank Vashisth Tiwari, Pranjal Aggarwal, Hamish Ivison, Anthony GX-Chen, and Daniel Fried for their useful and encouraging conversations about reward model discretization, and we thank Akhila Yerukola for her extensive assistance in refining our framing and preparing our manuscript.

\bibliographystyle{assets/plainnat}
\bibliography{example_paper}

\newpage
\appendix
\onecolumn
\section{Proof of Proposition \ref{prop:first} (Accuracy is the weighted sum of discriminative ability and specificity)}
\label{app:accuracy_oversensitivity_calculation}

Accuracy is
\begin{align*}
& \mathbb{E}_{a, b} [\mathbbm{1}[\sign(r_x(a) - r_x(b)) = \sign(u_x(a) - u_x(b))] \\
& = P\left(\sign(r_x(a) - r_x(b)) = \sign(u_x(a) - u_x(b))\right)\\
&= P\left[r_x(a) = r_x(b), u_x(a) = u_x(b)\right]\\
&\quad + P\left[r_x(a) > r_x(b), u_x(a) > u_x(b)\right]\\
&\quad + P\left[r_x(a) < r_x(b), u_x(a) < u_x(b)\right]
  \end{align*}
First term:
\begin{align*}
    &P\left[r_x(a) = r_x(b), u_x(a) = u_x(b)\right]\\
    &= \left(1 - P[r_x(a) \neq r_x(b) \mid u_x(a) = u_x(b) \right) \times P\left[u_x(a) = u_x(b)\right]
\end{align*}
Second and third terms:
\begin{align*}
&P\left[r_x(a) > r_x(b), u_x(a) > u_x(b)\right] + P\left[r_x(a) < r_x(b), u_x(a) < u_x(b)\right]\\
& = P\left[r_x(a) > r_x(b) \mid u_x(a) > u_x(b)\right] P(u_x(a) > u_x(b)) \hspace{0.5em} + P\left[r_x(a) < r_x(b) \mid u_x(a) < u_x(b)\right] P(u_x(a) < u_x(b)).
\end{align*}
Then, accuracy is
\begin{align*}
& \left(1 - P[r_x(a) \neq r_x(b) \mid u_x(a) = u_x(b) \right) \\
&\quad \times P\left[u_x(a) = u_x(b)\right]\tag*{\text{\emph{(specificity)}}}\\
& + P\left[r_x(a) > r_x(b) \mid u_x(a) > u_x(b)\right] P(u_x(a) > u_x(b))] \\
& + P\left[r_x(a) < r_x(b) \mid u_x(a) < u_x(b)\right] P(u_x(a) < u_x(b))] \tag*{\text{\emph{(discriminative ability)}}}\\ 
\end{align*}

\section{Proof of Proposition \ref{prop:discretized_reward_oversensitivity_under_uniform}
(The oversensitivity of a binary-discretized reward model is
\texorpdfstring{$\max(0, \tfrac{1}{2} - 2((\tau_x - \phi_x(u_x(a)))/s_x)^2)$}%
{max(0, 1/2 - 2((tau\_x - phi\_x(u\_x(a)))/s\_x)\^{}2)})}
\label{app:discretized_reward_oversensitivity_under_uniform_proof}

\paragraph{Proof} For sufficiently small $\epsilon$ ($\epsilon < s_x$), consider that
$|\text{disc}_x(r_x(a)) > \text{disc}_x(r_x(b))| \Longleftrightarrow |\text{disc}_x(r_x(a)) - \text{disc}_x(r_x(b))| > \epsilon$. Therefore, $P(|\text{disc}_x(r_x(a)) > \text{disc}_x(r_x(b))|  + \epsilon \mid u_x(a) = u_x(b))$ $=$ $P(\text{disc}_x(r_x(a)) \neq \text{disc}_x(r_x(b)) \mid u_x(a) = u_x(b))$

For our discretized (binary) reward to retain \textit{perfect discriminative ability}, this implies that there are exactly two unique values of utility for this task, meaning this task has a \textit{binary} notion of correctness. In this setting:
\begin{align*}
& P(\text{disc}_x(r_x(a)) \neq \text{disc}_x(r_x(b))) \quad = \quad P(\text{disc}_x(r_x(a)) > \text{disc}_x(r_x(b))) \quad + \quad P(\text{disc}_x(r_x(b)) > \text{disc}_x(r_x(a))).
\end{align*}
Conditioned on $u_x(a) = u_x(b)$, these two orderings are equally likely. We can compute the first and double it:

\begin{align*}
& P(\text{disc}_x(r_x(a)) > \text{disc}_x(r_x(b))) = P(r_x(a) > \tau_x \text{ and } r_x(b) \leq \tau_x) = \\
& P(\phi_x(u_x(a)) + \eta_x(a) > \tau_x  \quad \mkern9mu \text{ and } \mkern9mu \phi_x(u_x(b)) + \eta_x(b) \leq \tau_x) = \\
& P(\phi_x(u_x(a)) + \eta_x(a) > \tau_x) \mkern9mu \times \mkern9mu P(\phi_x(u_x(b)) + \eta_x(b) \leq \tau_x).
\end{align*}
The final line results from the fact that the events that $\phi_x(u_x(a)) + \eta_x(a) > \tau_x$ and $\phi_x(u_x(b)) + \eta_x(b) \leq \tau_x$ are independent, conditioned on the fact that $u_x(a) = u_x(b)$.

First term:
\begin{align*}
& P(\phi_x(u_x(a)) + \eta_x(a) > \tau_x) \\
& =P\bigl(\text{Unif}(-s_x/2, s_x/2) > \tau_x - \phi_x(u_x(a))\bigr) \\
& = P\bigl(\text{Unif}(-1/2, 1/2) > \text{\textit{distance}}\bigr) \\
& \quad\quad \textit{(where distance} := (\tau_x - \phi_x(u_x(a)))/s_x\text{\textit{)}} =\\
& = \begin{cases}
    1, & \text{if \textit{distance}} \leq -\frac{1}{2}\\
    0, & \text{if \textit{distance}} \geq \frac{1}{2}\\
    \frac{1}{2} -  \text{\textit{distance}}, & \text{otherwise.}
  \end{cases}
\end{align*}

Similarly, second term:
\begin{align*}
& P(\phi_x(u_x(b)) + \eta_x(b) \leq \tau_x) \\
& = P\bigl(\text{Unif}(-1/2, 1/2) \leq (\tau_x - \phi_x(u_x(b)))/s_x\bigr) \\
& = \begin{cases}
    0, & \text{if \textit{distance}} \leq -\frac{1}{2}\\
    1, & \text{if \textit{distance}}\geq \frac{1}{2}\\
    \text{\textit{distance}} + \frac{1}{2}, & \text{otherwise.}
  \end{cases}
\end{align*}

Given that $u_x(a) = u_x(b)$, we multiply these two terms and then double the result to account for both orderings:
\begin{align*}
& 2  P(\phi_x(u_x(a)) + \eta_x(a) > \tau_x)  \\
& \quad \quad \times P(\phi_x(u_x(b)) + \eta_x(b) \leq \tau_x) \\
& = 2\begin{cases}
    0, & \text{if \textit{distance}} \leq -\frac{1}{2}\\
    0, & \text{if \textit{distance}}\geq \frac{1}{2}\\
    1/4 - (distance)^2, & \text{otherwise.}
  \end{cases} \\
& = \max(0, 1/2 - 2(distance)^2) \\
& = \max\bigl(0, 1/2 - 2((\tau_x - \phi_x(u_x(a)))/s_x)^2\bigr)
\end{align*}

Then $\mathrm{Spec}_{\textup{disc}} = \min\bigl(1, 2((\tau_x - \phi_x(u_x(a)))/s_x)^2 - 1/2\bigr)$.

\section{Ablation of Number of Dropout Samples}
\label{sec:num_dropout_samples}

Section \ref{sec:discretization} introduces our algorithm for reward clustering, which uses MC dropout to estimate the predictive variance of a reward model. In \autoref{tab:t_ablation}, we show results from ablating  the number of samples taken via MC dropout for \textit{Skywork V1}. We find that the number of samples has surprisingly little impact on the intrinsic efficacy of our reward discretization method. We observe identical patterns with the other reward models used in this paper.

\begin{table}[h]
\centering
\small
\setlength{\tabcolsep}{4pt}
\begin{tabular}{ l | c c c | c c c}
\toprule
Method & \multicolumn{3}{c|}{\textit{Proposed} Metrics} & \multicolumn{3}{c}{Ties} \\
& Avg. & Discrim. & Specif. & Overall & Acc. & Correctness \\
\midrule
\multicolumn{7}{c}{\textit{Skywork V1}} \\
\midrule
\hspace{10pt} $T=3$ & 70.8 {\scriptsize ± 0.8} & 98.0 {\scriptsize ± 0.4} & 43.6 {\scriptsize ± 1.6} & 0.729 {\scriptsize ± 0.029} & 0.882 {\scriptsize ± 0.025} & 0.500 {\scriptsize ± 0.048} \\
$\bigstar$ $\boldsymbol{T=4}$ & 70.9 {\scriptsize ± 0.8} & 98.2 {\scriptsize ± 0.4} & 43.6 {\scriptsize ± 1.6} & 0.729 {\scriptsize ± 0.029} & 0.882 {\scriptsize ± 0.025} & 0.500 {\scriptsize ± 0.048} \\
\hspace{10pt} $T=8$ & 70.7 {\scriptsize ± 0.8} & 98.1 {\scriptsize ± 0.4} & 43.3 {\scriptsize ± 1.6} & 0.724 {\scriptsize ± 0.031} & 0.873 {\scriptsize ± 0.028} & 0.500 {\scriptsize ± 0.048} \\
\hspace{10pt} $T=12$ & 70.9 {\scriptsize ± 0.8} & 98.2 {\scriptsize ± 0.4} & 43.6 {\scriptsize ± 1.6} & 0.729 {\scriptsize ± 0.029} & 0.882 {\scriptsize ± 0.025} & 0.500 {\scriptsize ± 0.048} \\
\bottomrule
\end{tabular}
\caption{We find the effectiveness of reward clustering via MC dropout is robust to the number of dropout samples used. We report reward clustering performance for Skywork V1. $\bigstar$ indicates the default setting ($T=4$).}
\label{tab:t_ablation}
\end{table}

\section{Discretization still improves reward models if we use a Gaussian noise mode}
\label{app:gaussian_noise}

\subsection{A relaxed noise model: Gaussian-distributed rewards}
\label{sec:gaussian_noise_relaxation}

In Section~\ref{sec:definitions} we assumed bounded uniform noise
$\eta_x(a) \sim \text{Unif}(-s_x/2, s_x/2)$, leading to uniform-distributed rewards. These hard noise limits guaranteed perfect discriminative ability of the raw reward model;
without it, the analysis in
\S\ref{sec:discrim_and_oversensitivity}--\ref{sec:discretization_minimizes_oversensitivity}
no longer applies. We now relax this assumption to
\emph{Gaussian} noise. This means that the raw reward model can no longer be perfect. Here we assume that true utility is binary, as in \S\ref{sec:discrim_and_oversensitivity}, but these concepts readily generalize to multi-level utility functions.

\paragraph{Assumption.}
For a fixed prompt $x$, we model the reward as $r_x(a) = \phi_x(u_x(a)) + \eta_x(a)$, where $\phi_x(v) = s_x v + d_x$ and $\eta_x(a) \sim \mathcal{N}(0,\sigma_x^2)$. The per-prompt noise is homoskedastic: $\sigma_x$ and $s_x$ are shared across utility classes. We define the per-prompt \emph{signal-to-noise
ratio} as $\rho_x := s_x/\sigma_x$; this determines the reward model's accuracy.
For convenience, we drop the subscript $x$ and use $\Phi$ and $\varphi$ for the Gaussian CDF and density, respectively.

\subsection{Discriminative ability and oversensitivity under Gaussian noise}
\label{sec:gaussian_raw}

\begin{proposition}
\label{prop:gaussian_raw_discriminative}
Under the homoskedastic Gaussian model with binary utility,
\[
  D_{\textup{raw}}(\epsilon) = \bigl(r_x(a) - r_x(b) > \epsilon \mid u_x(a) > u_x(b)\bigr)
   = \Phi\left(\rho_x(1-\epsilon) / \sqrt{2}\right).
\]
\end{proposition}

\paragraph{Proof.}
For some $a, b$ where $u_x(a) > u_x(b)$, then $r_x(a) - r_x(b) = s_x + \eta_x(a) - \eta_x(b)$, where
$\eta_x(b)-\eta_x(a) \sim \mathcal{N}(0, 2\sigma_x^2)$. Thus
$D_{\textup{raw}}(\epsilon)
 = P\bigl(\eta_x(b) - \eta_x(a) < s_x - \epsilon\bigr)
 = \Phi\bigl((s_x-\epsilon)/(\sigma_x\sqrt 2)\bigr)$,
which equals $\Phi(\rho_x(1-\epsilon)/\sqrt 2)$.

As in \autoref{sec:discretization_minimizes_oversensitivity}, we measure reward in
\emph{utility units} to be able to directly compare between raw and discrete rewards. The noise within each utility class here is now 
$\eta_x/s_x \sim \mathcal{N}(0, 1/\rho_x^2)$.

\begin{proposition}
\label{prop:gaussian_raw_oversensitivity}
In utility units, for any $\epsilon \geq 0$,
\[
  \mathrm{Spec}_{\textup{raw}}(\epsilon) = P\bigl(|r_x(a) - r_x(b)|/s_x < \epsilon \bigm| u_x(a) = u_x(b)\bigr)
   = 2 \Phi(\epsilon \rho_x/\sqrt{2}) - 1.
\]
\end{proposition}

\paragraph{Proof.}
Conditional on $u_x(a)=u_x(b)$, $(r_x(a)-r_x(b))/s_x =
(\eta_x(a)-\eta_x(b))/s_x \sim \mathcal{N}(0, 2/\rho_x^2)$. Hence
$P(|\mathcal{N}(0,2/\rho_x^2)| < \epsilon) = P(\mathcal{N}(0,2/\rho_x^2) < \epsilon) - P(\mathcal{N}(0,2/\rho_x^2) < - \epsilon) = \Phi(\epsilon/\sqrt{2 / \rho_x^2}) - \Phi(-\epsilon/\sqrt{2 / \rho_x^2}) =$\\
$\Phi(\epsilon/\sqrt{2 / \rho_x^2}) - (1- \Phi(\epsilon/\sqrt{2 / \rho_x^2})) =$ $2\Phi(\epsilon \rho_x / \sqrt{2}) - 1$.

\subsection{Discretization preserves most discriminative ability and substantially reduces oversensitivity}
\label{sec:gaussian_disc}

\begin{proposition}
\label{prop:gaussian_disc_discriminative}
For binary utility, the discriminative ability of the discretized
reward, given two adjacent utility classes, is maximized at
$\tau_x := (\phi_x(u_x(a))+\phi_x(u_x(b)))/2$. Discriminative ability under this threshold is
\[
  D_{\textup{disc}}(\epsilon)  = P\bigl(\textup{disc}_x(r_x(a)) > \textup{disc}_x(r_x(b))
          \mid u_x(a) > u_x(b)\bigr)
   = \Phi(\rho_x/2)^2.
\]
\end{proposition}

\paragraph{Proof.}
Assume that $u_x(a)=1, u_x(b)=0$. Since the events $\{r_x(a) > \tau_x\}$ and $\{r_x(b) \leq \tau_x\}$ are independent:
\[
P\bigl(\textup{disc}_x(r_x(a)) > \textup{disc}_x(r_x(b)) \mid u_x(a) >  u_x(b)\bigr)
= \Phi\left(\tfrac{\phi_x(1)-\tau_x}{\sigma_x}\right)
  \Phi\left(\tfrac{\tau_x-\phi_x(0)}{\sigma_x}\right).
\]
We wish to find a value of $\tau_x$ that maximizes this probability. Let us denote $z := (\tau_x-\phi_x(0))/\sigma_x$ (this is the continuous analog of $\textit{distance}$ used in \autoref{app:discretized_reward_oversensitivity_under_uniform_proof})
. Then we can rewrite
$(\phi_x(1)-\tau_x)/\sigma_x$ as $\rho_x - z$. This in turn allows us to  rewrite $P\bigl(\textup{disc}_x(r_x(a)) > \textup{disc}_x(r_x(b)) \mid u_x(a) >  u_x(b)\bigr)$ as $\Phi(\rho_x-z)\Phi(z)$.

Now let
$f(z) := \Phi(\rho_x-z)\Phi(z)$. We can maximize this function by setting its derivative to zero (note that $\varphi$ refers to the derivative of $\Phi$):
\[
f'(z) = -\varphi(\rho_x-z)\Phi(z) + \Phi(\rho_x-z)\varphi(z) = 0 \Longleftrightarrow \tfrac{\varphi(z)}{\varphi(\rho_x-z))} = \tfrac{\Phi(z)}{\Phi(\rho_x-z)}\Longleftrightarrow
 \tfrac{\varphi(z)}{\Phi(z)} = \tfrac{\varphi(\rho_x-z)}{\Phi(\rho_x-z)} ,
\]
which holds at $z = \rho_x/2$. Then the optimizer is $\tau_x = \phi_x(0) +s_x/2 = (\phi_x(0) + \phi_x(1))/2$, and $D_{\textup{disc}}(\epsilon) = \Phi(\rho_x/2)^2$.

\begin{proposition}
\label{prop:gaussian_disc_oversensitivity}
At the threshold $\tau_x$ of
Proposition \ref{prop:gaussian_disc_discriminative}, for any
$0 \leq \epsilon < 1$ and binary utility:
\[
  \mathrm{Spec}_{\textup{disc}}(\epsilon)
   = 1 - P\bigl(|\textup{disc}_x(r_x(a)) - \textup{disc}_x(r_x(b))| > \epsilon
          \bigm| u_x(a) = u_x(b)\bigr)
   = 1 - 2\Phi(\rho_x/2)\Phi(-\rho_x/2).
\]
\end{proposition}

\paragraph{Proof.}
Since $\textup{disc}_x$ is binary-valued, then for $0\leq \epsilon < 1$, $|\textup{disc}_x(r_x(a)) - \textup{disc}_x(r_x(b))| > \epsilon \Longleftrightarrow \textup{disc}_x(r_x(a)) \neq \textup{disc}_x(r_x(b))$.

Let $z_u := (\tau_x - \phi_x(u))/\sigma_x$, representing the \textit{distance} from any given utility class. Conditional on
$u_x(a)=u_x(b)$, $\textup{disc}_x(r_x(a))$ and
$\textup{disc}_x(r_x(b))$ are independent. Then:
\[
P\bigl(\textup{disc}_x(r_x(a)) \neq \textup{disc}_x(r_x(b)) \mid u_x(a) = u_x(b)\bigr)
= 2\bigl(1 - \Phi(z_u)\bigr)\Phi(z_u)
= 2\Phi(-z_u)\Phi(z_u).
\]
Proposition \ref{prop:gaussian_disc_discriminative} shows discriminative ability is maximized at $\tau_x = \phi_x(0) +
s_x/2 = s_x/2 + d_x = (\phi_x(0) + \phi_x(1))/2$. Then, the distance of this threshold to $u_x(a) = 0$ is $z_0 = \rho_x/2$  and the distance to $u_x(a) = 1$ is $z_1 = -\rho_x/2$. Regardless of the utility class $u_x$, we then obtain the same oversensitivity: $2\Phi(\rho_x/2)\Phi(-\rho_x/2)$. Then $\mathrm{Spec}_{\textup{disc}}(\epsilon) = 1 - 2\Phi(\rho_x/2)\Phi(-\rho_x/2)$, taking the complement of oversensitivity.

\begin{remark}[the shape difference]
The oversensitivity of the discretized reward does not depend on $\epsilon$: it has a flat value of $2\Phi(\rho_x/2)\Phi(-\rho_x/2)$ across all $\epsilon \in [0,1)$. The raw oversensitivity depends on $\epsilon$: $2\Phi(-\epsilon\rho_x/\sqrt{2})$ starts at $1$ and
decays as $\epsilon$ increases. This shape difference is key for understanding why discretization is better (in
Theorem~\ref{thm:gaussian_success_condition}).
\end{remark}

\begin{remark}[choosing the threshold]
\label{rem:threshold_choice}
We use the midpoint discretization threshold $\tau_x = (\phi_x(0) + \phi_x(1))/2$ throughout. This threshold maximizes
discriminative ability (Proposition \ref{prop:gaussian_disc_discriminative}). It does not globally 
minimize oversensitivity, but rather it yields a favorable tradeoff, as we will show in \autoref{thm:gaussian_net_benefit}.
\end{remark}

\subsection{The discretization tradeoff}
\label{sec:gaussian_tradeoff}

Combining
Propositions~\ref{prop:gaussian_raw_discriminative}--\ref{prop:gaussian_disc_oversensitivity},
discretization reduces the discriminative ability from $\Phi(\rho_x/\sqrt{2})$ to $\Phi(\rho_x(1-\epsilon)/\sqrt{2})$, and discretization changes specificity from 
$2\Phi(\epsilon\rho_x/\sqrt{2})-1$
to
$1-2\Phi(\rho_x/2)\Phi(-\rho_x/2)$. The discretized reward 's discriminative ability and specificity are both  \emph{constant}  for $\epsilon \in [0,1)$. In contrast, the raw reward sees its discriminative ability
$\Phi(\rho_x(1-\epsilon)/\sqrt 2)$ decline while its specificity
$2\Phi(\epsilon\rho_x/\sqrt 2)-1$ rises from $0$ as $\epsilon$ grows.

\begin{theorem}
\label{thm:gaussian_success_condition}
For every signal-to-noise ratio $\rho_x > 0$ and every tolerance $\epsilon \in [0,
1/\sqrt{2}],$ $\mathrm{Spec}_{\textup{disc}}(\epsilon)
  > \mathrm{Spec}_{\textup{raw}}(\epsilon)$ (discretization improves the specificity of the reward model).
\end{theorem}

\paragraph{Proof.}
Propositions~\ref{prop:gaussian_raw_oversensitivity} and
\ref{prop:gaussian_disc_oversensitivity} give
$\mathrm{Spec}_{\textup{raw}}(\epsilon) = 2\Phi(\epsilon\rho_x/\sqrt 2) - 1$
and $\mathrm{Spec}_{\textup{disc}}(\epsilon) = 1 - 2\Phi(\rho_x/2)\Phi(-\rho_x/2)$.
Using $\Phi(\epsilon\rho_x/\sqrt 2) = 1 - \Phi(-\epsilon\rho_x/\sqrt 2)$, then
\begin{align*}
& \mathrm{Spec}_{\textup{disc}}(\epsilon) - \mathrm{Spec}_{\textup{raw}}(\epsilon)
  = 1 - 2\Phi(\rho_x/2)\Phi(-\rho_x/2) - 2(1 - \Phi(-\epsilon\rho_x/\sqrt 2)) + 1\\
& = 2\bigl[\Phi(-\epsilon\rho_x/\sqrt 2) - \Phi(\rho_x/2)\Phi(-\rho_x/2)\bigr],
\end{align*}
Because $\rho_x$ is positive, $\epsilon < 1 / \sqrt{2} \implies \epsilon / \sqrt{2} < 1/2 \implies \Phi(-\epsilon \rho_x/\sqrt{2}) > \Phi(-\rho_x/2)$. Because $\Phi(\rho_x/2) < 1$, $\Phi(-\epsilon \rho_x/\sqrt{2}) > \Phi(\rho_x/2) \Phi(-\rho_x/2)$, and then $\mathrm{Spec}_{\textup{disc}}(\epsilon) - \mathrm{Spec}_{\textup{raw}}(\epsilon) > 0$.

This means that for every $\epsilon$ below
$1/\sqrt{2}$ (in utility units) and at every signal-to-noise ratio, discretization is
strictly more specific than the raw reward. In practice, this means that the specificity improves as long as we declare two reward values as meaningfully different as long as their difference in rewards exceeds at least $\sim 70\%$ of the proportional difference in their true utility values. This covers most of the plausible range of $\epsilon$ that a practitioner might consider.

\paragraph{Weighing the two axes at a common tolerance.}
The success condition isolates specificity. To support the stronger
claim that discretization \emph{preserves most discriminative ability}
\emph{while} improving specificity, consider their average at a common tolerance $\epsilon$. The raw discriminative ability is
$D_{\textup{raw}}(\epsilon) = \Phi(\rho_x(1-\epsilon)/\sqrt 2)$
(Proposition \ref{prop:gaussian_raw_discriminative}); the discretized
discriminative ability is $D_{\textup{disc}}(\epsilon) = \Phi(\rho_x/2)^2$ (Proposition \ref{prop:gaussian_disc_discriminative}). As in \autoref{thm:uniform_sum}, we consider the single combined score 
$T_r(\epsilon) := (D_r(\epsilon) + \mathrm{Spec}_r(\epsilon))/2$.

\begin{proposition}
\label{prop:gaussian_net_benefit_minimizer}
Consider the averaged tradeoff between specificity and discrminative ability: $T(\epsilon) = (\mathrm{Spec}(\epsilon) + D(\epsilon))/2$. The gap between this value computed for a discretized reward model and the raw reward models is minimized at the \textit{worst-case} tolerance
$\epsilon = \min\{\tfrac12 + \tfrac{2\log 2}{\rho_x^2}, 1\}$.
\end{proposition}

\paragraph{Proof.}

Using $D_{\textup{disc}}(\epsilon) = \Phi(\rho_x/2)^2$ from Proposition \ref{prop:gaussian_disc_discriminative}, $\mathrm{Spec}_{\textup{disc}}(\epsilon) = 1 - 2\Phi(\rho_x/2)\Phi(-\rho_x/2)$ from Proposition \ref{prop:gaussian_disc_oversensitivity}, and
$\Phi(-\rho_x/2) = 1-\Phi(\rho_x/2)$, then
{\setlength{\abovedisplayskip}{3pt}\setlength{\abovedisplayshortskip}{3pt}\setlength{\belowdisplayskip}{3pt}\setlength{\belowdisplayshortskip}{3pt}
\[
T_{\textup{disc}} = (D_{\textup{disc}} + \mathrm{Spec}_{\textup{disc}})/2 \mkern9mu = \mkern9mu \bigl(\Phi(\rho_x/2)^2 \mkern9mu + \mkern9mu 1 - 2\Phi(\rho_x/2)(\Phi(-\rho_x/2))\bigr)/2 \mkern9mu = \mkern9mu\tfrac32 \Phi(\rho_x/2)^2 - \Phi(\rho_x/2) + \tfrac12
\]
}

Using $D_{\textup{raw}}(\epsilon) = \Phi\left(\rho_x(1-\epsilon) / \sqrt{2}\right)$ (Proposition \ref{prop:gaussian_raw_discriminative}) and $\mathrm{Spec}_{\textup{raw}}(\epsilon) =  2 \Phi(\epsilon \rho_x/\sqrt{2}) - 1$
(Proposition \ref{prop:gaussian_raw_oversensitivity}), 
{\setlength{\abovedisplayskip}{3pt}\setlength{\abovedisplayshortskip}{3pt}\setlength{\belowdisplayskip}{3pt}\setlength{\belowdisplayshortskip}{3pt}
\[
T_{\textup{raw}}(\epsilon)  = (D_{\textup{raw}} + \mathrm{Spec}_{\textup{raw}})/2 \mkern9mu = \mkern9mu \bigl(\Phi(\rho_x(1-\epsilon) / \sqrt{2}) + 2 \Phi(\epsilon \rho_x/\sqrt{2}) - 1\bigr)/2
\]
}

Denote $T_\Delta (\epsilon) = \mkern9mu T_{\textup{disc}}(\epsilon) - T_{\textup{raw}}(\epsilon) \mkern9mu = \mkern9mu \tfrac32 \Phi(\rho_x/2)^2 - \Phi(\rho_x/2) + \tfrac12 - (\bigl(\Phi(\rho_x(1-\epsilon) / \sqrt{2}) + 2 \Phi(\epsilon \rho_x/\sqrt{2}) - 1\bigr)/2)$.
First, we can ask where $T_\Delta$ is minimal. Setting the derivative to zero:
\begin{align*}
& T'_\Delta(\epsilon) = \tfrac{d}{d\epsilon} \bigl((-\Phi(\rho_x(1-\epsilon) / \sqrt{2}) - 2 \Phi(\epsilon \rho_x/\sqrt{2}) - 1)/2\bigr) = 0 \\
& \Longleftrightarrow \tfrac12\bigl(\tfrac{\rho_x}{\sqrt{2}} \varphi(\rho_x(1-\epsilon)/\sqrt{2}) - 2\tfrac{\rho_x}{\sqrt{2}} \varphi(\epsilon \rho_x / \sqrt{2})  \bigr) = 0 \\
& \Longleftrightarrow \tfrac{\rho_x}{2\sqrt{2}}\bigl(\varphi(\rho_x(1-\epsilon)/\sqrt{2}) - 2\varphi(\epsilon \rho_x / \sqrt{2})  \bigr) = 0 \\
& \implies 2 \varphi(\epsilon \rho_x / \sqrt{2})= \varphi(\rho_x(1-\epsilon)/\sqrt{2})\\
& \implies 2 \tfrac{1}{\sqrt{2 \pi}} \exp{\bigl(- \tfrac12 (\epsilon \rho/\sqrt{2})^2\bigr)} = \tfrac{1}{\sqrt{2 \pi}} \exp{\bigl(- \tfrac12 (\rho_x(1-\epsilon)/\sqrt{2})^2\bigr)}
\intertext{Taking the logarithm of both sides, we find a unique minimizer:}
& \implies \log 2 - \log(\sqrt{2\pi}) - \tfrac12 (\epsilon \rho/\sqrt{2})^2 = - \log(\sqrt{2\pi}) - \tfrac12 (\rho_x(1-\epsilon)/\sqrt{2})^2\\
& \implies \tfrac12 (\epsilon \rho/\sqrt{2})^2 - \tfrac12 (\rho_x(1-\epsilon)/\sqrt{2})^2 = \log 2 \\
& \implies \tfrac14 (2\epsilon-1)\rho_x^2 = \log 2 \implies \epsilon^* = \tfrac{2\log2}{\rho_x^2} + \tfrac12.
\end{align*}

We know the gradient is zero at this value is $\epsilon^*$. If $\epsilon^* \in [0, 1)$, then we have found an optimum on $[0, 1)$. Since this may not be the case (e.g. $\rho_x < 2 \sqrt{\log{2}}$), we can consider the sign of the gradient.
\begin{align*}
& \sign(T'_\Delta(\epsilon)) = \sign\Bigl(\tfrac{\rho_x}{2\sqrt{2}}\bigl(\varphi(\rho_x(1-\epsilon)/\sqrt{2}) - 2\varphi(\epsilon \rho_x / \sqrt{2})\bigr)\Bigr) \\
& = \sign\bigl(\varphi(\rho_x(1-\epsilon)/\sqrt{2}) - 2\varphi(\epsilon \rho_x / \sqrt{2})\bigr)
\intertext{Since both terms are positive, the inequality is preserved under the logarithm:}
& = \sign\bigl(\log\varphi(\rho_x(1-\epsilon)/\sqrt{2}) - \log(2\varphi(\epsilon \rho_x / \sqrt{2}))\bigr) \\
& = \sign\Bigl[\log\bigl(\tfrac{1}{\sqrt{2\pi}}\exp\bigl(-\tfrac12(\rho_x(1-\epsilon)/\sqrt{2})^2\bigr)\bigr) - \log\bigl(\tfrac{1}{\sqrt{2\pi}}\exp\bigl(-\tfrac12(\epsilon \rho_x / \sqrt{2})^2\bigr)\bigr) - \log 2\Bigr] \\
& = \sign\bigl(-\tfrac12(\rho_x(1-\epsilon)/\sqrt{2})^2 - (-\tfrac12(\epsilon \rho_x / \sqrt{2})^2) - \log 2\bigr) \\
& = \sign\bigl(\tfrac14 \rho_x^2(2\epsilon - 1) - \log 2\bigr)
\end{align*}
The gradient then changes sign at $\epsilon = \tfrac12 + \tfrac{2\log 2}{\rho_x^2}$; the gradient is negative for $\epsilon < \epsilon^*$ and positive for $\epsilon > \epsilon^*$. Therefore, we know that $\epsilon^*$ is the global minimum of $T_\Delta$, and we know that if $\epsilon > 1$, then $T_\Delta(1)$ is the minimum of $T_\Delta$ over $[0, 1)$. As we reduce $\epsilon \to 0$, the gap between  discretized and raw rewards grows further.

\begin{theorem}
\label{thm:gaussian_net_benefit}
 The specificity gain exceeds the discriminative cost: for every $\rho_x > 0$ and every tolerance $\epsilon \in [0,1): \quad T_{\textup{disc}}(\epsilon) - T_{\textup{raw}}(\epsilon) > 0$.
As $\rho_x \to \infty$, the net benefit approaches zero: when the raw reward itself approaches perfection, discretization is unnecessary.
\end{theorem}

\paragraph{Proof.}

Proposition \ref{prop:gaussian_net_benefit_minimizer} shows that $ T_{\textup{disc}}(\epsilon) - T_{\textup{raw}}(\epsilon)$ is minimized at $\epsilon = \min(1, \epsilon^*)$ where $\epsilon^* = \tfrac12 + \tfrac{2\log 2}{\rho_x^2}$.
If we can show that $T_\Delta\bigl(\min\{1,\epsilon^*\}\bigr) > 0$ for every $\rho_x > 0$; then $T_\Delta$ is positive for all values of $\epsilon \in [0, 1)$.

 Denote the worst-case value by $\Delta(\rho_x) := T_\Delta\bigl(\min\{1,\epsilon^*\}\bigr)$. We will divide our analysis by ranges of $\rho_x$. Our three cases are $0 < \rho_x \leq  2\sqrt{\log 2}$, $2\sqrt{\log 2} < \rho_x < 4.2$, and $\rho_x \geq 4.2$.

\paragraph{Case 1: $0 < \rho_x \le 2\sqrt{\log 2}$.}
Then, $\epsilon^* \geq 1$, so $T_\Delta$ is strictly decreasing on all of $[0,1)$: $T_\Delta(\epsilon) > T_\Delta(1)$. Since $T_\Delta$ is continuous at 1, it is therefore sufficient to show that  $T_\Delta(1) \geq 0$. 
\[
  T_\Delta(1) = \tfrac32\Phi(\rho_x/2)^2 - \Phi(\rho_x/2) - \Phi(\rho_x/\sqrt{2}) + \tfrac34 .
\]
Differentiating, and using $\varphi(\rho_x/\sqrt2) = \varphi(\rho_x/2)e^{-\rho_x^2/8}$,
\begin{align*}
  & \Delta'(\rho_x) = \tfrac32 \Phi(\tfrac{\rho_x}{2}) \varphi(\tfrac{\rho_x}{2}) - \tfrac12 \varphi(\tfrac{\rho_x}{2}) - \tfrac{\sqrt{2}}{2} \varphi(\tfrac{\rho_x}{\sqrt{2}}) \\
  & = \tfrac32 \Phi(\tfrac{\rho_x}{2}) \varphi(\tfrac{\rho_x}{2}) - \tfrac12 \varphi(\tfrac{\rho_x}{2}) - \tfrac{\sqrt{2}}{2} \bigl(-\tfrac{1}{\sqrt{2 \pi}} \exp{(-(\rho_x/\sqrt{2})^2/2)}\bigr) \\
  & = \tfrac32 \Phi(\tfrac{\rho_x}{2}) \varphi(\tfrac{\rho_x}{2}) - \tfrac12 \varphi(\tfrac{\rho_x}{2}) - \tfrac{\sqrt{2}}{2} \bigl(-\tfrac{1}{\sqrt{2 \pi}} \exp{(-(\rho_x/2)^2/2)} \exp{(-\rho_x^2/8)}\bigr) \\
  & = \tfrac32 \Phi(\tfrac{\rho_x}{2}) \varphi(\tfrac{\rho_x}{2}) - \tfrac12 \varphi(\tfrac{\rho_x}{2}) - \tfrac{\sqrt{2}}{2} \varphi(\tfrac{\rho_x}{2}) \exp{(-\rho_x^2/8)}\bigr) \\
& = \tfrac12 \varphi(\tfrac{\rho_x}{2}) \bigl(3\Phi(\tfrac{\rho_x}{2})  - 1 - \sqrt{2} \exp{(-\rho_x^2/8)}\bigr) \\
\end{align*}
$\tfrac12\varphi(\rho_x/2)$ is always positive. The other term $A(\rho_x) = 3\Phi(\rho_x/2) - 1 - \sqrt{2}e^{-\rho_x^2/8}$ is strictly increasing in $\rho_x$: $\tfrac{d}{d\rho_x} \bigl(A(\rho_x)\bigr) = $
$\tfrac32\varphi(\rho_x/2) + \tfrac{\sqrt2}{4}\rho_x e^{-\rho_x^2/8} > 0$ for $\rho_x > 0$. We know $A(0) = \tfrac12 - \sqrt{2} < 0$ and $A(2\sqrt{\log{2}}) = 3 \Phi(\sqrt{\log{2}}) - 2 > 0$, so its zero must lie between these two arguments of $A$. We can numerically find further that the zero (and, therefore, the minimizer of $\Delta(\rho_x)$) is between $\rho_x = 1.204$ and $\rho_x = 1.205$. At both endpoints here, where $\Delta(\rho_x) > 0.012$, and on this interval, $\tfrac12 \varphi(\rho_x/2) < 0.1665$, while $-0.0006 < A(\rho_x) < 0.0003$. A lower-bound value for $\Delta'(\rho_x)$ (most-negative possible slope) is $-10^{-4}$. Then, by the Mean Value Theorem, the minimum value of $\Delta(\rho_x)$ must be at least $0.012 - (1.205-1.204)\cdot10^{-4}$, so $\Delta(\rho_x)$ is always positive with a minimum value of at least 0.0119999.

\paragraph{Case 2: $\rho_x \geq 4.2$} , where this constant is chosen for reasons explained later in this section. 
Here $\epsilon^* < 1$ is the minimizer.
\[
  \Delta(\rho_x)
    = \tfrac32 \Phi(\rho_x/2)^2 - \Phi(\rho_x/2) + \tfrac12 - \bigl(\Phi(\rho_x(1-\epsilon) / \sqrt{2}) + 2 \Phi(\epsilon \rho_x/\sqrt{2}) - 1\bigr)/2.
\]

First, consider the limit as $\rho_x \to \infty$. Since $\lim_{x \to \infty} \Phi(x) = 1$, it is easy to see that $\lim_{\rho_x \to \infty} \Delta(\rho_x) =  \tfrac32 - 1 + \tfrac12 - 1 = 0$. Now we consider the finite case. For convenience, we will use the complement of the Gaussian cumulative distribution function  $\overline\Phi := 1 - \Phi$,
\begin{align*}
 & \Delta(\rho_x) = \tfrac32 (1 - \overline\Phi(\rho_x/2))^2 - (1- \overline\Phi(\rho_x/2)) + \tfrac12 - \bigl((1 - \overline\Phi(\rho_x(1-\epsilon) / \sqrt{2})) + 2 (1 - \overline\Phi(\epsilon \rho_x/\sqrt{2})) - 1\bigr)/2\\
&= \tfrac32\overline\Phi(\rho_x/2)^2 - 3\overline\Phi(\rho_x/2) + \tfrac32 - (1 - \overline\Phi(\rho_x/2)) + \tfrac12 - \bigl(2 - \overline\Phi(\rho_x(1-\epsilon)/\sqrt{2}) - 2\overline\Phi(\epsilon\rho_x/\sqrt{2})\bigr)/2\\
&= \tfrac32\overline\Phi(\rho_x/2)^2 - 2\overline\Phi(\rho_x/2) + \tfrac12\overline\Phi(\rho_x(1-\epsilon)/\sqrt{2}) + \overline\Phi(\epsilon\rho_x/\sqrt{2})
\intertext{Substituting $\tfrac12 + \tfrac{2\log 2}{\rho_x^2}$ for $\epsilon$:}
& = \tfrac32\overline\Phi(\rho_x/2)^2  - 2\overline\Phi(\rho_x/2) + \tfrac12\overline\Phi\Bigl(\tfrac{\rho_x}{2\sqrt2}-\tfrac{\sqrt2\log2}{\rho_x}\Bigr)
    + \overline\Phi\Bigl(\tfrac{\rho_x}{2\sqrt2}+\tfrac{\sqrt2\log2}{\rho_x}\Bigr).
\end{align*}

We can eliminate the two nonnegative terms $\overline\Phi(\tfrac{\rho_x}{2\sqrt2}+\tfrac{\sqrt2\log2}{\rho_x})$ and
$\tfrac32\overline\Phi(\rho_x/2)^2$:
\[
  \Delta(\rho_x) \ge
    \tfrac12\overline\Phi\Bigl(\tfrac{\rho_x}{2\sqrt2}-\tfrac{\sqrt2\log2}{\rho_x}\Bigr)
    - 2\overline\Phi(\rho_x/2).
\]

Then we need to show that $\tfrac12\overline\Phi\Bigl(\tfrac{\rho_x}{2\sqrt2}-\tfrac{\sqrt2\log2}{\rho_x}\Bigr) - 2\overline\Phi(\rho_x/2) > 0$, which is the same as 
\[
  \overline\Phi\bigl(\tfrac{\rho_x}{2\sqrt2}-\tfrac{\sqrt2\log2}{\rho_x}\bigr)/(\overline\Phi(\rho_x/2)) > 4.
\]

Recall that the Mills' ratio for the normal distribution requires that $t/(t^2+1) < \overline\Phi(t)/\varphi(t) < 1/t$.
We can apply the lower bound to the numerator and the upper bound to the denominator\\
$\overline\Phi\Bigl(\tfrac{\rho_x}{2\sqrt2}-\tfrac{\sqrt2\log2}{\rho_x}\Bigr) > \frac{\tfrac{\rho_x}{2\sqrt2}-\tfrac{\sqrt2\log2}{\rho_x}} {\bigl(\tfrac{\rho_x}{2\sqrt2}-\tfrac{\sqrt2\log2}{\rho_x}\bigr)^2 + 1} \varphi\Bigl(\tfrac{\rho_x}{2\sqrt2}-\tfrac{\sqrt2\log2}{\rho_x}\Bigr)$ and $\overline\Phi(\tfrac{\rho_x}2) < \tfrac{2}{\rho_x}\varphi(\rho_x/2)$.

Then, for the quantity that we want to show is greater than 4, we see a new lower bound:
\begin{align*}
\frac{\overline\Phi\bigl(\tfrac{\rho_x}{2\sqrt2}-\tfrac{\sqrt2\log2}{\rho_x}\bigr)}{\overline\Phi(\rho_x/2)} &> \frac{\rho_x\bigl(\tfrac{\rho_x}{2\sqrt2}-\tfrac{\sqrt2\log2}{\rho_x}\bigr)} {2\bigl[\bigl(\tfrac{\rho_x}{2\sqrt2}-\tfrac{\sqrt2\log2}{\rho_x}\bigr)^2 + 1\bigr]} \cdot \frac{\varphi\bigl(\tfrac{\rho_x}{2\sqrt2}-\tfrac{\sqrt2\log2}{\rho_x}\bigr)}{\varphi(\rho_x/2)}\\
&= \frac{\tfrac{\rho_x^2}{2\sqrt2}-\sqrt2\log2} {2\bigl[\bigl(\tfrac{\rho_x}{2\sqrt2}-\tfrac{\sqrt2\log2}{\rho_x}\bigr)^2 + 1\bigr]} \cdot \frac{\varphi\bigl(\tfrac{\rho_x}{2\sqrt2}-\tfrac{\sqrt2\log2}{\rho_x}\bigr)}{\varphi(\rho_x/2)}\\
&= \frac{\sqrt{2} (1 - \tfrac{4\log2}{\rho_x^2})}{1 + \tfrac{16 \log^2 2}{\rho_x^4} + \tfrac{8 - 8 \log 2}{\rho_x^2}} \cdot \frac{\varphi\bigl(\tfrac{\rho_x}{2\sqrt2}-\tfrac{\sqrt2\log2}{\rho_x}\bigr)}{\varphi(\rho_x/2)}
\end{align*}
Using $\varphi(t) = \tfrac{1}{\sqrt{2\pi}}e^{-t^2/2}$, we can simplify the second term here:
\begin{align*}
\frac{\varphi\bigl(\tfrac{\rho_x}{2\sqrt2}-\tfrac{\sqrt2\log2}{\rho_x}\bigr)}{\varphi(\rho_x/2)} &= \exp\Bigl(\tfrac12\Bigl[(\rho_x/2)^2 - \bigl(\tfrac{\rho_x}{2\sqrt2}-\tfrac{\sqrt2\log2}{\rho_x}\bigr)^2\Bigr]\Bigr) = \exp\Bigl(\tfrac12\Bigl[\tfrac{\rho_x^2}{4} - \tfrac{\rho_x^2}{8} - \tfrac{2\log^2 2}{\rho_x^2} + \log2\Bigr]\Bigr)\\
& = \exp\Bigl(\tfrac12\tfrac{\rho_x^2}{8}  - \tfrac12\tfrac{2 \log^2 2}{\rho_x^2} + \log2^{\tfrac12}\Bigr) = \sqrt2\exp\Bigl(\tfrac{\rho_x^2}{16} - \tfrac{\log^2 2}{\rho_x^2}\Bigr) %
.
\end{align*}
Now our lower bound is
{\setlength{\abovedisplayskip}{3pt}\setlength{\abovedisplayshortskip}{3pt}\setlength{\belowdisplayskip}{3pt}\setlength{\belowdisplayshortskip}{3pt}
\[
 \frac{\sqrt{2} (1 - \tfrac{4\log2}{\rho_x^2})}{1 + \tfrac{16 \log^2 2}{\rho_x^4} + \tfrac{8 - 8 \log 2}{\rho_x^2}} \cdot \sqrt2\exp\Bigl(\tfrac{\rho_x^2}{16} - \tfrac{\log^2 2}{\rho_x^2}\Bigr)
\]
}

Both terms increase as we $\rho_x$ goes from $2\sqrt{\log{2}} \to \infty$. Analytically, we can easily find values of $\rho_x$ where this lower bound exceeds 4. One such value is $\rho_x = 4.2$, where the lower bound is $4.24$. Therefore, we know $\Delta(\rho_x) > 0$ for all $\rho_x \geq 4.2$, and we will use this $\rho_x = 4.2$ as the floor of this case.

Therefore, we must consider one final case, where $2\sqrt{\log{2}} < \rho_x < 4.2$. This case fails our (loose) lower bound. Since this is a bounded subset of $\mathbb{R}$, we can use interval arithmetic \citep{moore1966interval}.

\paragraph{Case 3: $2\sqrt{\log{2}} < \rho_x < 4.2$.}
We will substitute $\epsilon = \tfrac12 + \tfrac{2\log 2}{\rho_x^2}$ into our definition of $\Delta(\rho_x)$:
\begin{align*}
  \Delta(\rho_x)
    & = \tfrac32 \Phi(\rho_x/2)^2 - \Phi(\rho_x/2) + \tfrac12 - \bigl(\Phi(\rho_x(1-\epsilon) / \sqrt{2}) + 2 \Phi(\epsilon \rho_x/\sqrt{2}) - 1\bigr)/2\\
    & = \tfrac32 \Phi(\rho_x/2)^2 - \Phi(\rho_x/2) + \tfrac12 - \tfrac12 \Phi\bigl(\rho_x(1- (\tfrac12 + \tfrac{2\log 2}{\rho_x^2})  ) / \sqrt{2}\bigr) - \Phi\bigl((\tfrac12 + \tfrac{2\log 2}{\rho_x^2})\rho_x/\sqrt{2}\bigr) + \tfrac12 \\
    & = \tfrac32 \Phi(\rho_x/2)^2 - \Phi(\rho_x/2) - \tfrac12 \Phi\bigl(\rho_x(1- (\tfrac12 + \tfrac{2\log 2}{\rho_x^2})  ) / \sqrt{2}\bigr) - \Phi\bigl((\tfrac12 + \tfrac{2\log 2}{\rho_x^2})\rho_x/\sqrt{2}\bigr) + 1\\
    & = \tfrac32 \Phi(\rho_x/2)^2 - \Phi(\rho_x/2) - \tfrac12 \Phi\bigl(\tfrac{\rho_x}2 - \tfrac{2\log 2}{\rho_x})  / \sqrt{2}\bigr) - \Phi\bigl((\tfrac{\rho_x}2 + \tfrac{2\log 2}{\rho_x})/\sqrt{2}\bigr) + 1 \\
    & = \underbrace{\tfrac32(\Phi(\rho_x/2) - \tfrac13)^2}_{(t_1)} - \underbrace{\tfrac12 \Phi\bigl(\tfrac{\rho_x}2 - \tfrac{2\log 2}{\rho_x})  / \sqrt{2}\bigr)}_{(t_2)} - \underbrace{\Phi\bigl((\tfrac{\rho_x}2 + \tfrac{2\log 2}{\rho_x})/\sqrt{2}\bigr) + \tfrac56}_{(t_3)}
\end{align*}

As $\rho_x$ increases, $t_1$ rises, while $t_2$ and $t_3$ fall. For any interval $[a, b]$, $a$ minimizes $t_1$ while $b$ minimizes $t_1$ and $t_2$. Then $\min_{\rho \in [a,b]} \bigl(t_1(\rho) + t_2(\rho) + t_3(\rho)\bigr) > \min_{\rho \in [a,b]}t_1(\rho) + \min_{\rho \in [a,b]}t_2(\rho) + \min_{\rho \in [a,b]}t_3(\rho) = t_1(a) + t_2(b) + t_3(b)$. 

Then, we can separate $\rho_x$ into two separate values $\rho_a$ and $\rho_b$ to define a strict lower bound for $\Delta(\rho_x)$:
\[
 \tfrac32(\Phi(\underline{\mathbf{\rho_b}}/2) - \tfrac13)^2 - \tfrac12 \Phi\bigl(\underline{\mathbf{\rho_a}}/2 - 2\log 2/\underline{\mathbf{\rho_a}})  / \sqrt{2}\bigr) - \Phi\bigl((\underline{\mathbf{\rho_a}}/2 + 2\log 2/\underline{\mathbf{\rho_a}})/\sqrt{2}\bigr) + \tfrac56
\]

We can then analytically find an arbitrary covering of intervals over $[2\sqrt{\log{2}}, 4.2)$ with positive lower bounds:

\noindent
\begin{tabular}{l c}
\toprule
Interval Endpoints & Lower Bound on $\Delta(\rho)$ \\
\midrule
$(2\sqrt{\log 2},\mkern9mu 1.8]$ & 0.0059\\
$[1.8,\mkern9mu 2.0]$ & 0.0053\\
$[2.0,\mkern9mu 2.2]$ & 0.0159\\
$[2.2,\mkern9mu 2.4]$ & 0.0256\\
$[2.4,\mkern9mu 2.8]$ & 0.0104\\
$[2.8,\mkern9mu 3.4]$ & 0.0065\\
$[3.4,\mkern9mu 4.2]$ & 0.0094\\
\bottomrule
\end{tabular}

Aggregating over all these intervals, we have $\min_{\epsilon \in [2\sqrt{\log{2}}, 4.2)} \Delta(\rho) > 0.0053 > 0$, so $T_\Delta\bigl(\min\{1,\epsilon^*\}\bigr) > 0$ over $\rho \in (2\sqrt{\log{2}}), 4.2)$. Earlier, we showed that $T_\Delta\bigl(\min\{1,\epsilon^*\}\bigr) > 0$ over $\rho \in (0, 2\sqrt{\log2}]$ and $\rho \in (4.2, \infty)$.

Putting this all together, $T_\Delta(\rho)$ is guaranteed to be positive over all $\rho \in (0, \infty)$. Therefore, $T_{\textup{disc}} > T_{\textup{raw}}$ for all $\rho_x > 0$ and all $\epsilon \in [0,1)$.

\newpage

\end{document}